%% file: main.tex
\documentclass{article}

     \usepackage[final, nonatbib]{neurips_2020}

\usepackage[round]{natbib} %round parenthesis
\usepackage{placeins} %floatbarrier
\usepackage{adjustbox} %resizable tables
\usepackage{amsmath}
\usepackage{xcolor}
\usepackage{float} %it positions tables in appendix
\usepackage{algorithmicx}
\usepackage[ruled]{algorithm}
\usepackage{algpseudocode}
\usepackage{subcaption}
\usepackage{bbm} %for indicator function
\usepackage{listings} %for Python code

\usepackage[utf8]{inputenc} % allow utf-8 input
\usepackage[T1]{fontenc}    % use 8-bit T1 fonts
\usepackage{hyperref}       % hyperlinks
\usepackage{url}            % simple URL typesetting
\usepackage{booktabs}       % professional-quality tables
\usepackage{amsfonts}       % blackboard math symbols
\usepackage{nicefrac}       % compact symbols for 1/2, etc.
\usepackage{microtype}      % microtypography

\title{Self-Supervised Relational Reasoning for Representation Learning}

\author{
  Massimiliano Patacchiola\\
  School of Informatics\\
  University of Edinburgh\\
  \texttt{mpatacch@ed.ac.uk} \\
  \And
  Amos Storkey\\
  School of Informatics\\
  University of Edinburgh\\
  \texttt{a.storkey@ed.ac.uk} \\
}

\begin{document}

\maketitle

\begin{abstract}
  In self-supervised learning, a system is tasked with achieving a surrogate objective by defining alternative targets on a set of unlabeled data. The aim is to build useful representations that can be used in downstream tasks, without costly manual annotation. In this work, we propose a novel self-supervised formulation of relational reasoning that allows a learner to bootstrap a signal from information implicit in unlabeled data. 
  Training a relation head to discriminate how entities relate to themselves (\emph{intra-reasoning}) and other entities (\emph{inter-reasoning}), results in rich and descriptive representations in the underlying neural network backbone, which can be used in downstream tasks such as classification and image retrieval. We evaluate the proposed method following a rigorous experimental procedure, using standard datasets, protocols, and backbones. Self-supervised relational reasoning outperforms the best competitor in all conditions by an average 14\% in accuracy, and the most recent state-of-the-art model by 3\%. We link the effectiveness of the method to the maximization of a Bernoulli log-likelihood, which can be considered as a proxy for maximizing the mutual information, resulting in a more efficient objective with respect to the commonly used contrastive losses.
\end{abstract}

\section{Introduction}

Learning useful representations from unlabeled data can substantially reduce dependence on costly manual annotation, which is a major limitation in modern deep learning. Toward this end, one solution is to develop learners able to self-generate a supervisory signal exploiting implicit information, an approach known as self-supervised learning \citep{schmidhuber1987evolutionary,schmidhuber1990making}.
Humans and animals are naturally equipped with the ability to learn via an intrinsic signal, but how machines can build similar abilities has been material for debate \citep{lake2017building}.
A common approach consists of defining a surrogate task (\emph{pretext}) which can be solved by learning generalizable representations, then use those representations in \emph{downstream} tasks, e.g. classification and image retrieval \citep{jing2020self}.

A key factor in self-supervised human learning is the acquisition of new knowledge by relating entities, whose positive effects are well established in studies of adult learning \citep{gentner2005relational, goldwater2018relational}. Developmental studies have shown something similar in children, who can build complex taxonomic names when they have the opportunity to compare objects \citep{gentner1999comparison, namy2002making}. Comparison allows the learner to neglect irrelevant perceptual features and focus on non-obvious properties. Here, we argue that it is possible to exploit a similar mechanism in self-supervised machine learning via relational reasoning.

The relational reasoning paradigm is based on a key design principle: the use of a relation network as a learnable function to quantify the relationships between a set of \emph{objects}.
Starting from this principle, we propose a new formulation of relational reasoning which can be used as a pretext task to build useful representations in a neural network backbone, by training the relation head on unlabeled data.
Differently from the canonical relational approach, which focuses on relations between \emph{objects in the same scene} \citep{santoro2017simple}, we focus on relations between \emph{views of the same object} (intra-reasoning) and relations between \emph{different objects in different scenes} (inter-reasoning), in doing so we allow the learner to acquire both intra-class and inter-class knowledge without the need of labeled data.

We evaluate our method following a rigorous experimental methodology, since comparing self-supervised learning methods can be problematic \citep{kolesnikov2019revisiting, musgrave2020metric}. Gains may be largely due to the backbone and learning schedule used, rather than the self-supervised component. To neutralize these effects we provide a benchmark environment where all methods are compared using standard datasets (CIFAR-10, CIFAR-100, CIFAR-100-20, STL-10, tiny-ImageNet, SlimageNet), evaluation protocol \citep{kolesnikov2019revisiting}, learning schedule, and backbones (both shallow and deep). Results show that our method largely outperforms the best competitor in all conditions by an average $14\%$ accuracy and the most recent state-of-the-art method by $3\%$.

\emph{Main contributions}: 1)~we propose a novel algorithm based on relational reasoning for the self-supervised learning of visual representations, 2)~we show its effectiveness on standard benchmarks with an in-depth experimental analysis, outperforming concurrent state-of-the-art methods (code released with an open-source license\footnote{\url{https://github.com/mpatacchiola/self-supervised-relational-reasoning}}), and 3)~we highlight how the maximization of a Bernoulli log-likelihood in concert with a relation module, results in more effective and efficient objective functions with respect to the commonly used contrastive losses.

\subsection{Overview}

Following the terminology used in the self-supervised literature \citep{jing2020self} we consider relational reasoning as a \emph{pretext} task for learning useful representations in the underlying neural network backbone. Once the joint system (backbone + relation head) has been trained, the relation head is discarded, and the backbone used in \emph{downstream} tasks (e.g. classification, image retrieval). To achieve this goal we provide a new formulation of relational reasoning. The \emph{canonical formulation} defines it as the process of learning the ways in which entities are connected, using this knowledge to accomplish higher-order goals \citep{santoro2017simple, santoro2018relational}. The \emph{proposed formulation} defines it as the process of learning the ways entities relate to themselves (intra-reasoning) and to other entities (inter-reasoning), using this knowledge to accomplish downstream goals.

Consider a set of objects $\mathcal{O} = \{ o_1, \ldots , o_N \}$, the canonical approach is \emph{within-scene}, meaning that all the elements in $\mathcal{O}$ belong to the same scene (e.g. fruits from a basket). The within-scene approach is not very useful in our case. Ideally, we would like our learner to be able to differentiate between objects taken from every possible scene. Therefore first we define \emph{between-scenes} reasoning: the task of relating objects from different scenes (e.g. fruits from different baskets).

Starting from the between-scenes setting, consider the case where the learner is tasked with discriminating if two objects $\{o_i, o_j\} \sim \mathcal{O}$ belong to the same category $\{o_i, o_j\} \rightarrow$~\emph{same}, or to a different one $\{o_i, o_j\} \rightarrow$~\emph{different}. Often a single attribute is informative enough to solve the task. For instance, in the pair $\{\text{apple}_i, \text{orange}_j \}$ the color alone is a strong predictor of the class, it follows that the learner does not need to pay attention to other features, this results in poor representations. 

To solve the issue we alter the object $o_i$ via random augmentations $\mathcal{A}(o_i)$ (e.g. geometric transformation, color distortion) making between-scenes reasoning more complicated. The color of an orange can be randomly changed, or the shape resized, such that it is much more difficult to discriminate it from an apple. In this challenging setting, the learner is forced to take account of the correlation between a wider set of features (e.g. color, size, texture, etc.).

However, it is not possible to create pairs of similar and dissimilar objects when labels are not given. To overcome the problem we bootstrap a supervisory signal directly from the (unlabeled) data, and we do so by introducing \emph{intra-reasoning} and \emph{inter-reasoning}. Intra-reasoning consists of sampling two random augmentations of the same object $\{\mathcal{A}(o_{i}), \mathcal{A}(o_{i}) \} \rightarrow$ \emph{same} (positive pair), whereas inter-reasoning consists of coupling two random objects $\{\mathcal{A}(o_{i}), \mathcal{A}(o_{\setminus i}) \} \rightarrow$ \emph{different} (negative pair). This is like coupling different views of the same apple to build the positive pair, and coupling an apple with a random fruit to build the negative pair.
In this work we show that it is possible to train a relation module via intra-reasoning and inter-reasoning, with the aim of learning useful representations.

\section{Previous work}\label{sec:previous_work}

\textbf{Relational reasoning.} In the last decades there have been entire sub-fields interested in relational learning: e.g. reinforcement learning \citep{dvzeroski2001relational} and statistics \citep{koller2007introduction}. However, only recently the relational paradigm has gained traction in the deep learning community with applications in question answering \citep{santoro2017simple, raposo2017discovering}, graphs \citep{battaglia2018relational}, sequential streams \citep{santoro2018relational}, deep reinforcement learning \citep{zambaldi2019deep}, few-shot learning \citep{sung2018learning}, and object detection \citep{hu2018relation}.
Our work differentiate from previous one in several ways: (i) previous work is based on labeled data, while we use relational reasoning on unlabeled data; (ii) previous work has focused on within-scene relations, here we focus on relations between different views of the same object (intra-reasoning) and between different objects in different scenes (inter-reasoning); (iii) in previous work training the relation head was the main goal, here is a pretext task for learning useful representations in the underlying backbone.

\textbf{Solving pretext tasks.} There has been a substantial effort in defining self-supervised pretext tasks which can be solved only if generalizable representations have been learned. Examples are: predicting the augmentation applied to a patch \citep{dosovitskiy2014discriminative}, predicting the relative location of patches \citep{doersch2015unsupervised}, solving Jigsaw puzzles \citep{noroozi2016unsupervised}, learning to count \citep{noroozi2017representation}, spotting artifacts \citep{jenni2018self}, predicting image rotations \citep{gidaris2018unsupervised}, or image channels \citep{zhang2017split}, generating color version of grayscale images \citep{zhang2016colorful, larsson2016learning}, and generating missing patches \citep{pathak2016context}.

\textbf{Metric learning.} The aim of metric learning~\citep{bromley1994signature} is to use a distance metric to bring closer representations of similar inputs (positives), while moving away representations of dissimilar inputs (negatives). Commonly used losses are the contrastive loss~\citep{hadsell2006dimensionality}, the
triplet loss~\citep{weinberger2006distance}, the Noise-Constrative Estimation (NCE, \citealt{gutmann2010noise}), the margin~\citep{schroff2015facenet} and magnet~\citep{rippel2016metric} losses.
At a first glance relational reasoning and metric learning may seem related, however they are fundamentally different: 
(i) metric learning explicitly aims at organizing representations by similarity, self-supervised relational reasoning aims at learning a relation measure and, as a byproduct, learning useful representations; (ii) metric learning directly applies a distance metric over the representations, relational reasoning collects representations into a set, aggregates them, then estimates relations; (iii) the relational score is not a distance metric (see Section~\ref{ssec:relation_module}) but rather a learnable (probabilistic) similarity measure.

\textbf{Contrastive learning.} Metric learning methods based on contrastive loss and NCE are often referred to as contrastive learning methods.
Contrastive learning via NCE has recently obtained the state of the art in self-supervised learning. However, one limiting factor is that NCE relies on a large quantity of negatives, which are difficult to obtain in mini-batch stochastic optimization. Recent work has used a memory bank to dynamically store negatives during training \citep{wu2018unsupervised}, followed by a plethora of other methods \citep{he2019momentum,tian2019contrastive, misra2019self,zhuang2019local}. However, a memory bank has several issues, it introduces additional overhead and a considerable memory footprint. SimCLR \citep{chen2020simple} tries to circumvent the problem by mining negatives in-batch, but this requires specialized optimizers to stabilize the training at scale.
We compare relational reasoning and constrastive learning in Section~\ref{ssec:inputs_augmentation} and Section~\ref{sec:discussion}.

\textbf{Pseudo-labeling.} Self-supervision can be achieved providing pseudo-labels to the learner, which are then used for standard supervised learning. A way to obtain pseudo-labels is to use the model itself, picking up the class which has the maximum predicted probability \citep{lee2013pseudo, sohn2020fixmatch}.
A neural network ensemble can also be used to provide the labels \citep{gupta2020unsupervised}.
In DeepCluster \citep{caron2018deep}, pseudo-labels are produced by running a k-means clustering algorithm, which can be forced to induce equipartition \citep{asano2020self}. Recent studies have shown that pseudo-labeling is not competitive against other methods \citep{oliver2018realistic}, since they are often prone to degenerate solutions with points assigned to the same label (or cluster).

\textbf{InfoMax.} A recent line of work has investigated the use of mutual information for unsupervised and self-supervised representation learning, following the InfoMax principle \citep{linsker1988self}.
Mutual information is often maximized at different scales (global and local) on single views (Deep InfoMax, \citealt{hjelm2018learning}), multi-views \citep{bachman2019learning, ji2019invariant}, or sequentially \citep{oord2018representation}.
Those methods are often strongly dependent on the choice of feature extractor architecture \citep{tschannen2020mutual}.

\begin{figure}[t]
 \centering
 \includegraphics[width=1.0\textwidth]{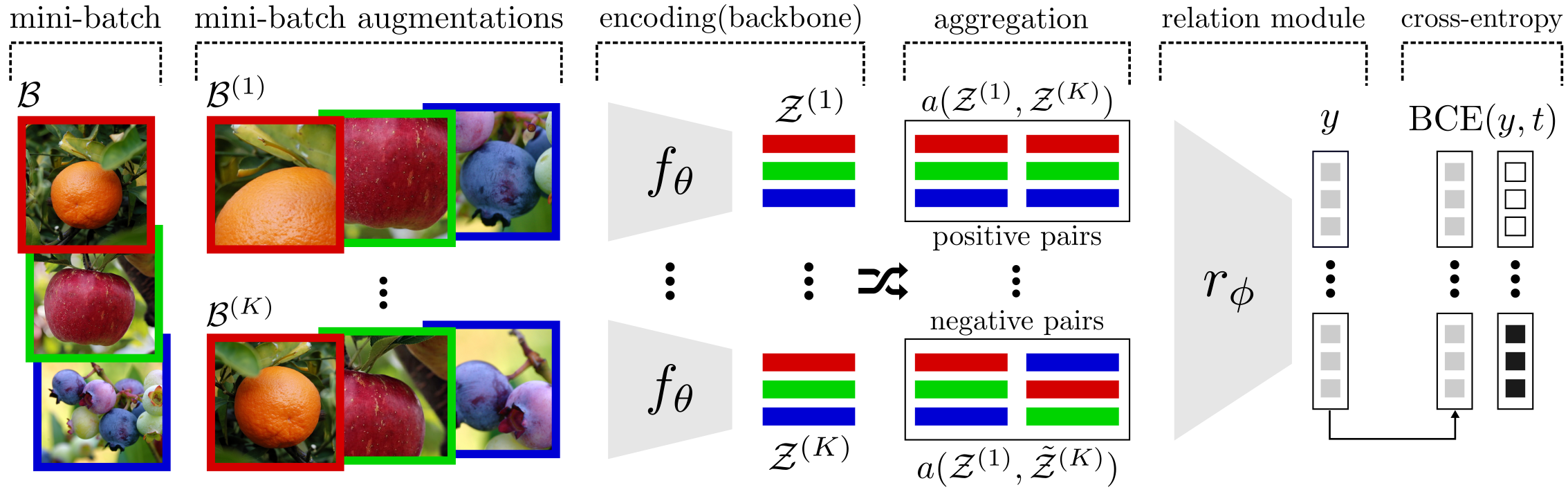}
 \caption{Overview of the proposed method. The mini-batch $\mathcal{B}$ is augmented $K$ times (e.g. via random flip and crop-resize) and passed through a neural network backbone $f_{\theta}$ to produce the representations $\mathcal{Z}^{(1)}, \ldots, \mathcal{Z}^{(K)}$. An aggregation function $a$ joins positives (representations of the same images) and negatives (randomly paired representations) through a commutative operator. The relation module $r_{\phi}$ estimates the relational score $y$, which must be 1 for positives and 0 for negatives. The model is optimized minimizing the binary cross-entropy (BCE) between prediction and target $t$.}
 \label{fig:model_overview}
\end{figure}

\section{Description of the method}\label{sec:method}

Consider an unlabeled dataset $\mathcal{D}=\{ \mathbf{x}_n \}_{n=1}^{N}$ and a non-linear function $f_{\theta}(\cdot)$ parameterized by a vector of learnable weights $\boldsymbol{\theta}$, modeled as a neural network (backbone). A forward pass generates a vector $f_{\theta}(\mathbf{x}_{n})=\mathbf{z}_n$ (representation), which can be collected in a set $\mathcal{Z} = \{ \mathbf{z}_{n} \}_{n=1}^{N}$.
The notation $\mathcal{A}(\mathbf{x}_{n})$ is used to express the probability distribution of instances generated by applying stochastic data augmentation to $\mathbf{x}_{n}$, while $\mathbf{x}_{n}^{(i)} \sim \mathcal{A}(\mathbf{x}_{n})$ is the $i$-th sample from this distribution (a particular augmented version of the input instance), and $\mathcal{D}^{(i)}=\{ \mathbf{x}^{(i)}_n \}_{n=1}^{N}$ the $i$-th set of random augmentations over all instances.
Likewise $\mathbf{z}_{n}^{(i)}=f_{\theta}(\mathbf{x}_{n}^{(i)})$ is grouped in $\mathcal{Z}^{(i)} = \{ \mathbf{z}^{(i)}_{n} \}_{n=1}^{N}$. Let $K$ indicate the total number of augmentations $\mathcal{D}^{(1)},\ldots,\mathcal{D}^{(K)}$ and their representations $\mathcal{Z}^{(1)},\ldots,\mathcal{Z}^{(K)}$.
Now, let us define a relation module $r_{\phi}(\cdot)$, as a non-linear function approximator parameterized by $\boldsymbol{\phi}$, which takes as input a pair of aggregated representations and returns a relation score $y$. Indicating with $a(\cdot, \cdot)$ an aggregation function and with $\mathcal{L}(y, t)$ the loss between the score and a target value $t$, the complete learning objective can be specified as
\begin{equation}\label{equation:overall}
    \underset{\boldsymbol{\theta}, \boldsymbol{\phi}}{\text{argmin}}
    \
    \sum_{n=1}^{N} \sum_{i=1}^{K} \sum_{j=1}^{K} 
    \underbrace{\mathcal{L}\Big( r_{\phi} \big( a(\mathbf{z}_{n}^{(i)}, \mathbf{z}_{n}^{(j)}) \big) , t=1 \Big)}_{\text{intra-reasoning}}
    +
    \underbrace{\mathcal{L}\Big( r_{\phi} \big( a(\mathbf{z}_{n}^{(i)}, \mathbf{z}_{\setminus n}^{(j)}) \big) , t=0 \Big)}_{\text{inter-reasoning}},
    \ \
    \text{with}
    \ \
    \mathbf{z}_{n} = f_{\theta}(\mathbf{x}_{n}),
\end{equation}
where $\setminus n$ is an index randomly sampled from $\{1, \ldots, N \} \setminus \{ n \}$. 
In practice \eqref{equation:overall} can be framed as a standard binary classification problem (see Section~\ref{ssec:definition_loss}), and minimized by stochastic gradient descent sampling a mini-batch $\mathcal{B} \sim \mathcal{D}$ with pairs built by repeatedly applying $K$ augmentations to $\mathcal{B}$. Positives can be obtained pairing two encodings of the same input (intra-reasoning term), and negatives by randomly coupling representations of different inputs (inter-reasoning term), relying on the assumption that in common settings this yields a very low probability of false negatives.
An overview of the model is given in Figure~\ref{fig:model_overview} and the pseudo-code in Appendix~\ref{appendix:code}.

\textbf{Mutual information.} Following the recent work of \cite{boudiaf2020metric} we can interpret~\eqref{equation:overall} in terms of \emph{mutual information}. Let us define the random variables $Z|X$ and $T|Z$, representing embeddings and targets. Now consider the generative view of mutual information 
\begin{equation}\label{equation:mutual-information}
I(Z; T) = H(Z) - H(Z|T).
\end{equation}
Intra-reasoning is a tightening factor which can be expressed as a bound over the conditional entropy $H(Z|T)$. Inter-reasoning is a scattering factor which can be linked to the entropy of the representations $H(Z)$. In other words, each representation is pushed towards a positive neighborhood (intra-reasoning) and repelled from a complementary set of negatives (inter-reasoning).
Under this interpretation~\eqref{equation:overall} can be considered as a proxy for maximizing Equation~\eqref{equation:mutual-information}. We refer the reader to~\cite{boudiaf2020metric} for a more detailed analysis.

\subsection{Inputs augmentation}\label{ssec:inputs_augmentation}

Given a random mini-batch of $M$ input instances $\mathcal{B} \sim \mathcal{D}$, recursively apply data augmentation $K$ times $\mathcal{B}^{(1)}, \dots, \mathcal{B}^{(K)}$ then propagate through $f_{\theta}$ with a forward pass, to generate the corresponding representations $\mathcal{Z}^{(1)}, \dots, \mathcal{Z}^{(K)}$.
Representations are coupled across augmentations to generate positive and negative tuples
\begin{equation}
 \forall i, j \in \{1, \dots, K\}
 \quad
 \big(\underbrace{ \mathcal{Z}^{(i)}, \mathcal{Z}^{(j)} }_{\text{positives}}\big)
 \quad
 \text{and}
 \quad
 \big(\underbrace{ \mathcal{Z}^{(i)}, \tilde{\mathcal{Z}}^{(j)} }_{\text{negatives}}\big),
\end{equation}
where $\tilde{\mathcal{Z}}$ indicates random assignment of each representation $\mathbf{z}^{(i)}_{n}$ to a different element $\mathbf{z}^{(j)}_{\setminus n}$. In practice, we discard identical pairs (identity mapping is learned across augmentations) and take just one of the symmetrical tuples $(\mathbf{z}^{(i)}, \mathbf{z}^{(j)})$ and $(\mathbf{z}^{(j)}, \mathbf{z}^{(i)})$ (the aggregation function ensures commutation, see Section~\ref{ssec:aggregation_funciton}).
If a certain amount of data in $\mathcal{D}$ is labeled (semi-supervised setting), then positive pairs include representations of different augmented inputs belonging to the same category.

\textbf{Computational cost.} Having defined $M$ as the number of inputs in the mini-batch $\mathcal{B}$, and $K$ as the number of augmentations, the total number of pairs $P$ (positive and negative) is given by
\begin{equation}
P = M (K^{2}- K).
\end{equation}

The number of comparisons $P$ scales quadratically with the number of augmentations $K$, and linearly with the size of the mini-batch $M$; whereas in recent constrastive learning methods \citep{chen2020simple}, they scale as $P = (MK)^{2}$, which is quadratic in both augmentations and mini-batch size. 

\textbf{Augmentation strategy.}
Here, we consider the particular case where the input instances are color images. Following previous work \citep{chen2020simple} we focus on two augmentations: random crop-resize and color distortion. Crop-resize enforces comparisons between views: global-to-global, global-to-local, and local-to-local. Since augmentations are sampled from the same color distribution, the color alone may suffice to distinguish positives and negatives. Color distortion enforces color-invariant encodings and neutralizes learning shortcuts. Additional details about the augmentations used in this work are reported in Section~\ref{sec:experiments} and Appendix~\ref{appendix:augmentations}.

\subsection{Aggregation function}\label{ssec:aggregation_funciton}

Relation networks operate over sets. To avoid a combinatorial explosion due to an increasing cardinality, a commutative aggregation function is applied. Given  $f_{\theta}(\mathbf{x}_i)=\mathbf{z}_i$ and $f_{\theta}(\mathbf{x}_j)=\mathbf{z}_j$, there are different possible choices for the aggregation function
\begin{equation}
    a_{\text{sum}}(\mathbf{z}_i, \mathbf{z}_j) = \mathbf{z}_i + \mathbf{z}_j, \quad
    a_{\text{max}}(\mathbf{z}_i, \mathbf{z}_j) = \text{max}(\mathbf{z}_i, \mathbf{z}_j),
    \quad
     a_{\text{cat}}(\mathbf{z}_i, \mathbf{z}_j) = \big( \mathbf{z}_i, \mathbf{z}_j \big),
\end{equation}
where sum and max are applied elementwise. Concatenation $a_{\text{cat}}$ is not commutative, but it has been previously used when the cardinality is small \citep{hu2018relation, sung2018learning}, like in our case.

\subsection{Relation module}\label{ssec:relation_module}

The relation module is a function $r_{\phi}(\cdot)$ parameterized by a vector of learnable weights $\boldsymbol{\phi}$, modeled as a multi-layer perceptron (MLP). Given a pair of representations $\mathbf{z}_i$ and $\mathbf{z}_j$, the module takes as input the aggregated pair and produce a scalar $y$ (relation score)
\begin{equation}
r_{\phi} \big( a(\mathbf{z}_i, \mathbf{z}_j) \big)=y.
\end{equation}

The relational score respects two properties: (i) $r(a(\mathbf{z}_i, \mathbf{z}_j)) \in [0,1]$; (ii) $r(a(\mathbf{z}_i, \mathbf{z}_j))=r(a(\mathbf{z}_j, \mathbf{z}_i))$.
It is crucial to not misinterpret the relational score for a pairwise distance metric. Given a set of input vectors $\{ \mathbf{v}_i, \mathbf{v}_j, \mathbf{v}_k \}$ the distance metric $d(\cdot, \cdot)$ respects four properties: (i) $d(\mathbf{v}_i, \mathbf{v}_j) \geq 0$; (ii) $d(\mathbf{v}_i, \mathbf{v}_j) = 0 \leftrightarrow \mathbf{v}_i=\mathbf{v}_j$; (iii) $d(\mathbf{v}_i,\mathbf{v}_j)=d(\mathbf{v}_j,\mathbf{v}_i)$; (iv) $d(\mathbf{v}_i,\mathbf{v}_k) \leq d(\mathbf{v}_i,\mathbf{v}_j)+d(\mathbf{v}_j,\mathbf{v}_k)$.  Note that the relational score does not satisfies all the conditions of a distance metric and therefore \emph{the relational score is not a distance metric}, but rather a probabilistic estimate (see Section~\ref{ssec:definition_loss}).

\subsection{Definition of the loss}\label{ssec:definition_loss}

The learning objective~\eqref{equation:overall} can be framed as a binary classification problem over the $P$ representation pairs. Under this interpretation, the relation score $y$ represents a probabilistic estimate of representation membership, which can be induced through a sigmoid activation function.
It follows that the objective reduces to the maximization of a Bernoulli log-likelihood, or similarly, the minimization of a binary cross-entropy loss
\begin{equation}
\mathcal{L}(\mathbf{y}, \mathbf{t}, \gamma) = \frac{1}{P} \sum_{i=1}^{P} - w_i \Big[ t_i \cdot \log y_i + (1 - t_i) \cdot \log (1 - y_i) \Big],
\end{equation}
with target $t_i=1$ for positives and $t_i=0$ for negatives. The optional weight $w_i$ is a scaling factor
\begin{equation}
w_i = \frac{1}{2} \Big[ (1 - t_i) \cdot y_i + t_i \cdot (1 - y_i) \Big]^{\gamma},
\end{equation}
where $\gamma \geq 0$ defines how sharp the weight should be. This factor gives more importance to uncertain estimations and it is also known as the focal loss \citep{lin2017focal}.
Note that, a binary estimator has been previously used in the context of correlation minimization for independent component analysis \citep{brakel2017learning} and in information maximization for representation learning \citep{hjelm2018learning}. \cite{hjelm2018learning} did not find any major benefit in using a binary loss (the Jensen-Shannon estimator), but similarly to us they observed a low sensitivity to the number of negative samples, outperforming NCE as the number of negatives became smaller (see Section~\ref{sec:discussion} for a discussion).

\section{Experiments}\label{sec:experiments}

Evaluating self-supervised methods is problematic because of substantial inconsistency in the way methods have been compared \citep{kolesnikov2019revisiting, musgrave2020metric}. We provide a standardized environment implemented in Pytorch using standard datasets (CIFAR-10, CIFAR-100, CIFAR-100-20, STL-10, tiny-ImageNet, SlimageNet), different backbones (shallow and deep), same learning schedule (epochs), and well know evaluation protocols \citep{kolesnikov2019revisiting}. In most conditions our method show superior performance.

\textbf{Implementation.} Hyperparameters (relation learner): mini-batch of 64 images ($K=16$ for ResNet-32 on tiny-ImageNet, $K=25$ for ResNet-34 on STL-10, $K=32$ for the rest), Adam optimizer with learning rate $10^{-3}$, binary cross-entropy loss with focal factor ($\gamma=2$). Relation module: MLP with 256 hidden units (batch-norm + leaky-ReLU) and a single output unit (sigmoid). Aggregation: we used concatenation as it showed to be more effective (see Appenidx~\ref{appendix:additional_ablations}, Table~\ref{tab:aggregation_comparison}). Augmentations: horizontal flip (50\% chance), random crop-resize, conversion to grayscale (20\% chance), and color jitter (80\% chance). Backbones: Conv-4, ResNet-8/32/56 and ResNet-34 \citep{he2016deep}.
Baselines: DeepCluster \citep{caron2018deep}, RotationNet \citep{gidaris2018unsupervised}, Deep InfoMax \citep{hjelm2018learning}, and SimCLR \citep{chen2020simple}. Those are recent (hard) baselines, with SimCLR being the current state-of-the-art in self-supervised learning. As upper bound we include the performance of a fully supervised learner (it has access to the labels), and as lower bound a network initialized with random weights, evaluated training only the linear classifier.
All results are the average over three random seeds. Additional details in Appendix~\ref{appendix:implementation_details}.

\textbf{Linear evaluation.} We follow the linear evaluation protocol defined by \cite{kolesnikov2019revisiting} training the backbone for 200 epochs using the \emph{unlabeled} training set, and then training for 100 epochs a linear classifier on top of the backbone features (without backpropagation in the backbone weights). The accuracy of this classifier on the test set is considered as the final metric to asses the quality of the representations. Our method largely outperforms other baselines with an accuracy of 46.2\% (CIFAR-100) and 30.5\% (tiny-Imagenet), which is an improvement of +4.0\% and +4.7\% over the best competitor (SimCLR), see Table~\ref{tab:results}. Best results are also obtained with the Conv-4 backbone on all datasets. Only in CIFAR-10/ResNet-32 SimCLR is doing better, with a score of 77\% against 75\% of our method, see  Appendix~\ref{appendix:additional_linear_evaluation}. In the appendix we report the results on the challenging SlimageNet dataset used in few-shot learning~\citep{antoniou2020defining}: 160 low-resolution images for each one of the 1000 classes in ImageNet. On SlimageNet our method has the highest accuracy (15.8\%, $K=16$), being better than RotationNet (7.2\%) and SimCLR (14.3\%).

\textbf{Domain transfer.}  We evaluate the performance of all methods in transfer learning by training on the unlabeled CIFAR-10 with linear evaluation on the labeled CIFAR-100 (and viceversa). Our method outperforms once again all the others in every condition. In particular, it is very effective in generalizing from a simple dataset (CIFAR-10) to a complex one (CIFAR-100), obtaining an accuracy of 41.5\%, which is a gain of +5.3\% over SimCLR and +7.5\% over the supervised baseline (with linear transfer). For results see Table~\ref{tab:results} and Appendix~\ref{appendix:additional_domain_transfer}.

\textbf{Grain.} Different methods produce different representations, some may be better on datasets with a small amount of labels (coarse-grained), others may be better on datasets with a large amount of labels (fine-grained). To investigate the granularity of the representations we train on unlabeled CIFAR-100, then perform linear evaluation using the 100 labels (fine grained; e.g. apple, fox, bee, etc) and the 20 super-labels (coarse grained; e.g. fruits, mammals, insects, etc). Also in this case our method is superior in all conditions with an accuracy of 52.4\% on CIFAR-100-20, see Table~\ref{tab:results} and Appendix~\ref{appendix:additional_grain}. In comparison, the method does better in the fine-grained case, indicating that it is well suited for datasets with a large amount of classes.

\textbf{Finetuning.} We used the STL-10 dataset \citep{coates2011analysis} which provides a set of unlabeled data coming from a similar but different distribution from the labeled data. Methods have been trained for 300 epochs on the unlabeled set (100K images), finetuned for 20 epochs on the labeled set (5K images), and finally evaluated on the test set (8K images). We used a mini-batch of 64 with $K=25$ and a ResNet-34. Implementation details are reported in Appendix~\ref{appendix:implementation_finetuning}. Results in Table~\ref{tab:results} show that our method obtains the highest accuracy: 89.67\% (best seed 90.04\%). Moreover a wider comparison reported in Appendix~\ref{appendix:additional_finetuning} shows that the method outperforms strong supervised baselines and the previous self-supervised state-of-the-art (88.80\%, \citealp{ji2019invariant}).

\begin{table}[t!]
 \caption{Comparison on various benchmarks. Mean accuracy (percentage) and standard deviation over three runs (ResNet-32). Best results in bold. \textbf{Linear Evaluation:} training on unlabeled data and linear evaluation on labeled data. \textbf{Domain Transfer:} training on unlabeled CIFAR-10 and linear evaluation on labeled CIFAR-100 (10$\rightarrow$100), and viceversa (100$\rightarrow$10). \textbf{Grain:} training on unlabeled CIFAR-100, linear evaluation on coarse-grained CIFAR-100-20 (20 super-classes). \textbf{Finetune:} training on the unlabeled set of STL-10, finetuning on the labeled set (ResNet-34).}
 \label{tab:results}
 \begin{adjustbox}{width=\columnwidth,center}
  \centering
  \begin{tabular}{lcccccc}
    \toprule
     & \multicolumn{2}{c}{\textbf{Linear Evaluation}} & \multicolumn{2}{c}{\textbf{Domain Transfer}} &
     \multicolumn{1}{c}{\textbf{Grain}} &
     \multicolumn{1}{c}{\textbf{Finetune}}\\
    \cmidrule[0.1pt](r){2-3} \cmidrule[0.1pt](r){4-5} \cmidrule[0.1pt](l){6-6} \cmidrule[0.1pt](l){7-7}
    \textbf{Method} &
    \textbf{\small{CIFAR-100}} & \textbf{\small{tiny-ImgNet}} &
    \textbf{\small{10$\rightarrow$100}} & \textbf{\small{100$\rightarrow$10}} &
    \textbf{\small{CIFAR-100-20}} & \textbf{\small{STL-10}} \\
    \midrule
    Supervised (upper bound) & 
    65.32$\pm$\small{0.22} & 50.09$\pm$\small{0.32}  & 33.98$\pm$\small{0.71} & 71.01$\pm$\small{0.44} & 76.35$\pm$\small{0.57} & 69.82$\pm$\small{3.36} \\
    Random Weights (lower bound) & 
    7.65$\pm$\small{0.44} & 3.24$\pm$\small{0.43} & 7.65$\pm$\small{0.44} & 27.47$\pm$\small{0.83} & 16.56$\pm$\small{0.48} & n/a \\
    \cmidrule(l){1-7}
    DeepCluster \citep{caron2018deep} & 
    20.44$\pm$\small{0.80} & 11.64$\pm$\small{0.21} & 18.37$\pm$\small{0.41} & 43.39$\pm$\small{1.84} & 29.49$\pm$\small{1.36} & 73.37$\pm$\small{0.55} \\
    RotationNet \citep{gidaris2018unsupervised} & 
    29.02$\pm$\small{0.18} & 14.73$\pm$\small{0.48} & 27.02$\pm$\small{0.20} & 52.22$\pm$\small{0.70} & 40.45$\pm$\small{0.39} & 83.29$\pm$\small{0.44} \\
    Deep InfoMax \citep{hjelm2018learning} & 
    24.07$\pm$\small{0.05} & 17.51$\pm$\small{0.15} & 23.73$\pm$\small{0.04} & 45.05$\pm$\small{0.24} & 33.92$\pm$\small{0.34} & 76.03$\pm$\small{0.37} \\
    SimCLR \citep{chen2020simple} & 
    42.13$\pm$\small{0.35} & 25.79$\pm$\small{0.35} & 36.20$\pm$\small{0.16} & 65.59$\pm$\small{0.76} & 51.88$\pm$\small{0.48} & 89.31$\pm$\small{0.14} \\  
    \emph{Relational Reasoning} (ours)  & \textbf{46.17$\pm$\small{0.17}} & \textbf{30.54$\pm$\small{0.42}} & \textbf{41.50$\pm$\small{0.35}} & \textbf{67.81$\pm$\small{0.42}} & \textbf{52.44$\pm$\small{0.47}} & \textbf{89.67$\pm$\small{0.33}} \\
    \bottomrule
  \end{tabular}
 \end{adjustbox}
\end{table}

\begin{figure}[t!]
    \begin{subfigure}[t]{0.333\textwidth}
        \includegraphics[width=1.0\textwidth, trim={0.25cm 0.8cm 0.25cm 0.25cm}, clip]{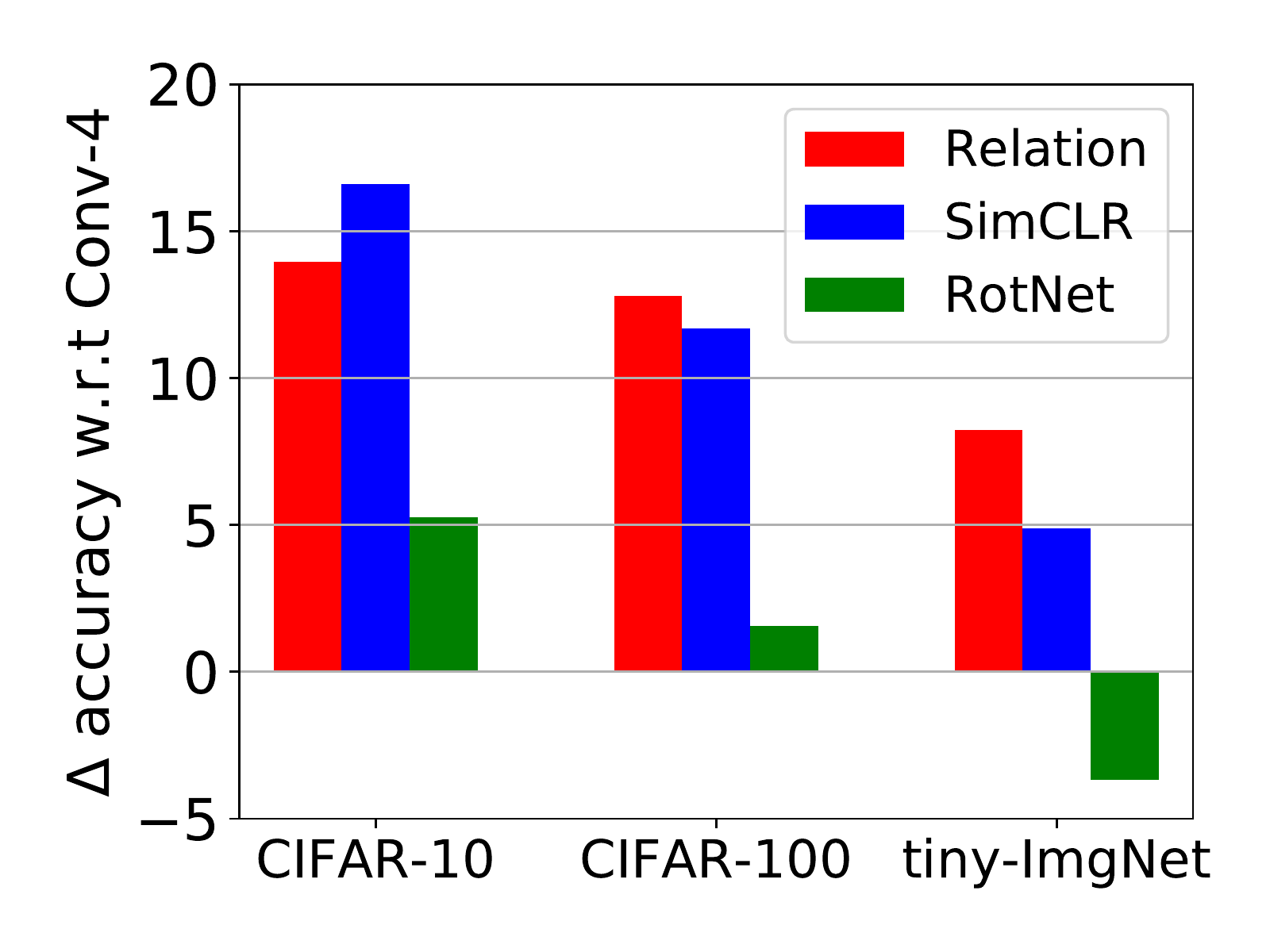}
        \caption{}
        \label{fig:results-architectures}
    \end{subfigure}%
    \begin{subfigure}[t]{0.333\textwidth}
        \centering
        \includegraphics[width=1.0\textwidth, trim={0.25cm 0.8cm 0.25cm 0.25cm}, clip]{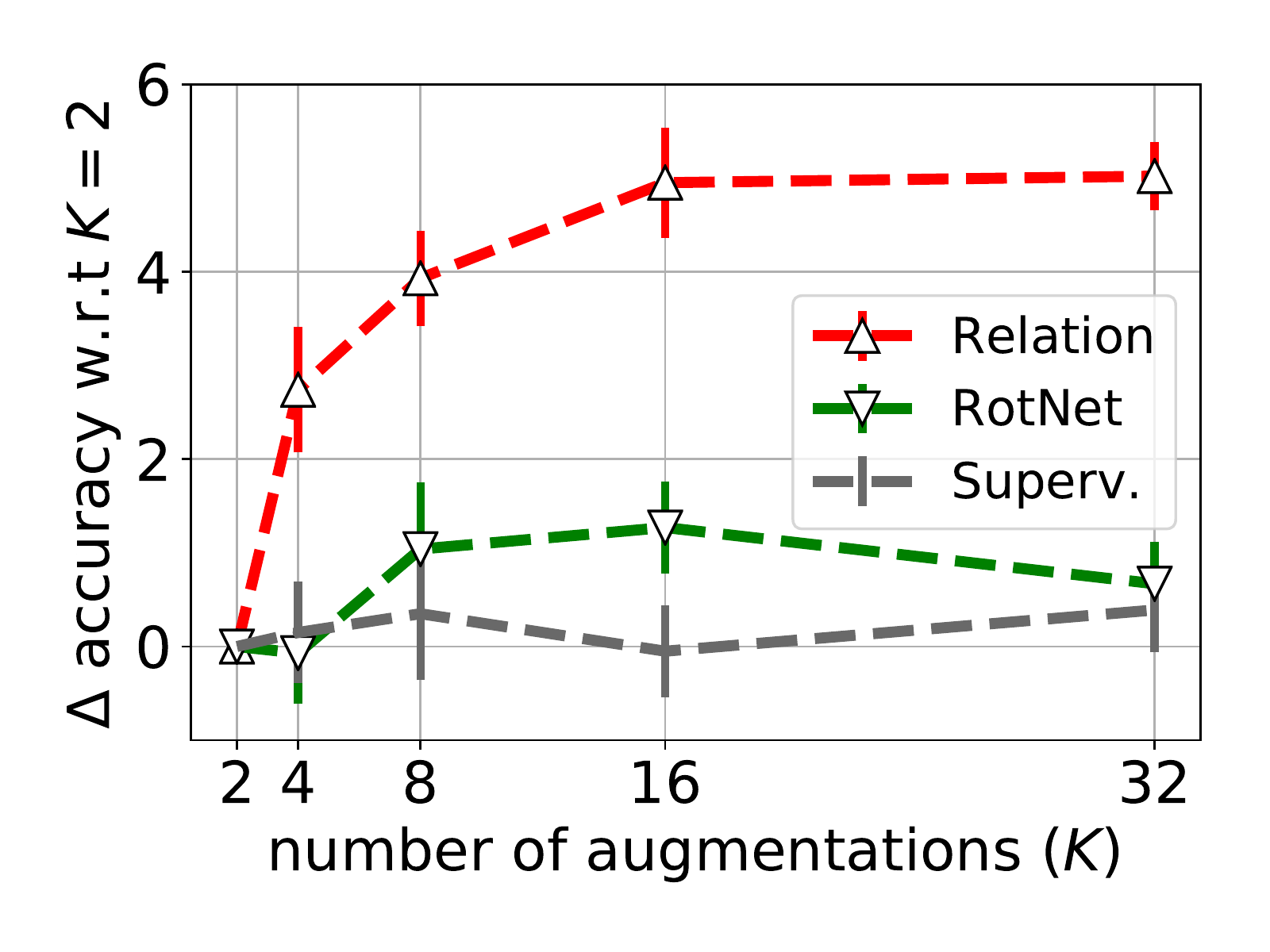}
        \caption{}
        \label{fig:results-augmentations}
    \end{subfigure}
    \begin{subfigure}[t]{0.333\textwidth}
        \centering
        \includegraphics[width=1.0\textwidth, trim={0.25cm 0.8cm 0.25cm 0.25cm}, clip]{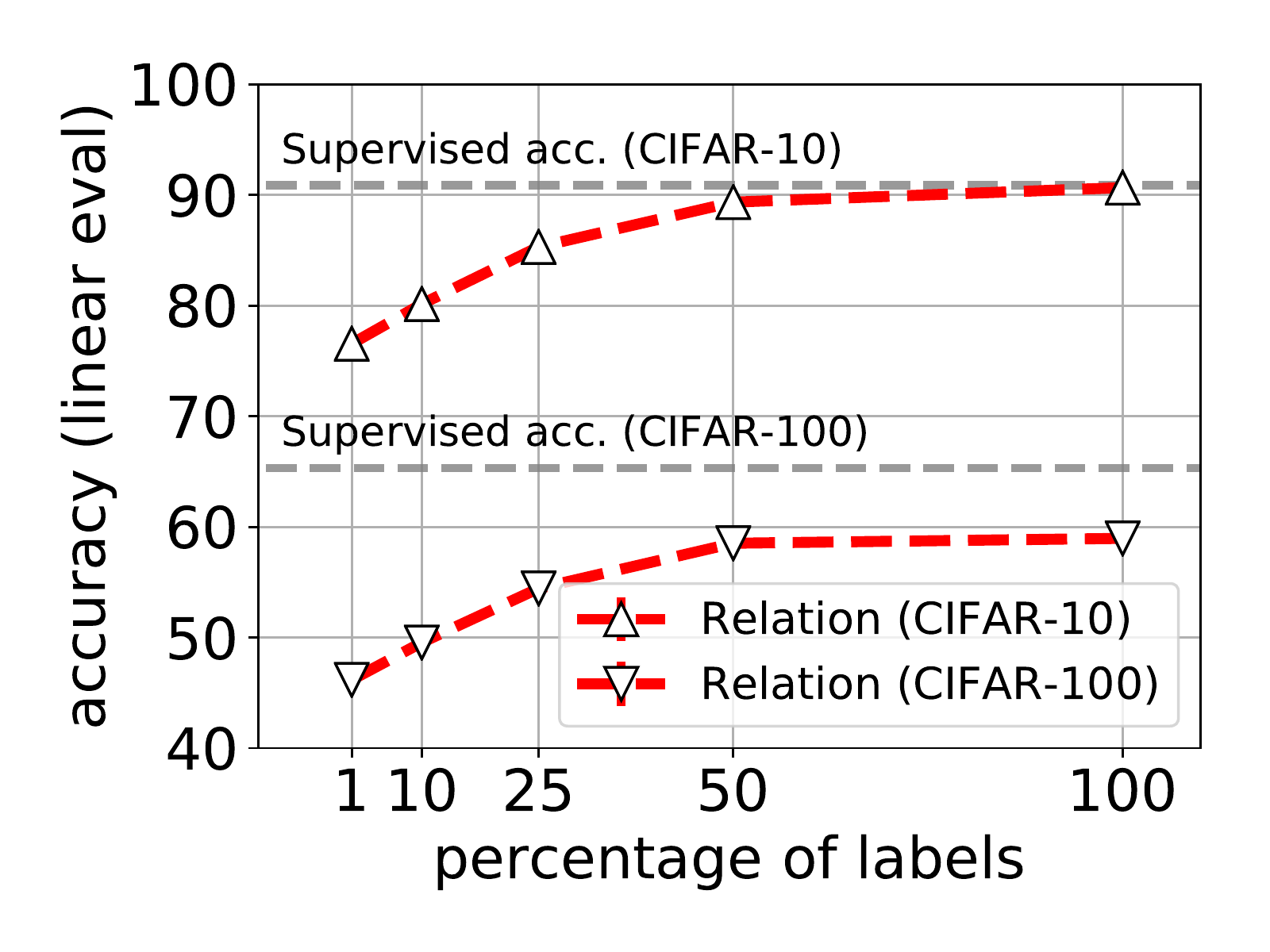}
        \caption{}
        \label{fig:results-semisupervised}
    \end{subfigure}
    \caption{(\subref{fig:results-architectures})~Difference in accuracy using the deeper backbone (Conv4$\rightarrow$ResNet-32, linear evaluation). As the complexity of the dataset raises our method performs increasingly better than the others. (\subref{fig:results-augmentations})~Correlation between validation accuracy (3 seeds, Conv-4, CIFAR-10) and number of mini-batch augmentations. Only in our method the accuracy is positively correlated with the number of augmentations. (\subref{fig:results-semisupervised})~Semi-supervised accuracy with an increasing percentage of labels (ResNet-32).}
    \label{fig:results}
\end{figure}

\textbf{Depth of the backbone.} In Appendix~\ref{appendix:additional_backbones} we report an extensive comparison on four backbones of increasing depth: Conv-4, ResNet-8, ResNet-32, and ResNet-56. We tested the three best methods (RotationNet, SimCLR, and Relational Reasoning) on CIFAR-10/100 linear evaluation, grain, and domain transfer for a total of 24 conditions. Results show that our method has the highest accuracy on 21 of those conditions, with SimCLR performing better on CIFAR-10 linear evaluation with ResNet backbones.
A distilled version of those results is reported in Figure~\ref{fig:results-architectures}. The figure shows the gain in accuracy from using a ResNet-32 instead of a Conv-4 backbone for datasets of increasing complexity (10, 100, and 200 classes). As the complexity of the dataset raises our method performs increasingly better than the others. The relative gain against SimCLR gets larger: $-2.6\%$ (CIFAR-10), $+1.1\%$ (CIFAR-100), $+3.3\%$ (tiny-ImageNet). The relative gain against RotationNet is even more evident: $+8.7\%$, $+11.2\%$, $+11.9\%$.

\textbf{Additional experiments.}
Figure~\ref{fig:results-augmentations} and Appendix~\ref{appendix:number_augmentations} show the difference in accuracy between $K=2$ and $K \in \{4, 8, 16, 32\}$ mini-batch augmentations for a fixed mini-batch size. There is a clear positive correlation between the number of augmentations and the performance of our model, while the same does not hold for a self-supervised algorithm (RotationNet) and the supervised baseline.
Figure~\ref{fig:results-semisupervised} and Appendix~\ref{appendix:semisupervised} show the accuracy obtained via linear evaluation in the semi-supervised setting, when the number of available labels is gradually increased (0\%, 1\%, 10\%, 25\%, 50\%, 100\%), in both CIFAR-10 and CIFAR-100 (ResNet-32). The accuracy is positively correlated with the proportion of labels available, approaching the supervised upper bound when 100\% of labels are available (supervised case).

\textbf{Ablations.} In Appendix~\ref{appendix:additional_ablations} we report the results of ablation studies on the aggregation function and relation head. We compare four aggregation functions: sum, mean, maximum, and concatenation. Results show that concatenation and maximum are respectively the most and less effective functions. Concatenation may favor backpropagation improving the quality of the representations, as supported by similar results in previous work \citep{sung2018learning}. Ablations of the relation head have followed two directions: (i) removing the head, and (ii) replacing the relation module with an encoder. In the first condition we removed the head and replace it with a simple dot product between representation pairs (BCE-focal loss). In the second condition we followed an approach similar to SimCLR~\citep{chen2020simple}, replacing the relation head with an encoder and applying the dot product to representations at the higher level (BCE-focal loss). The second condition differs from SimCLR for the loss type (BCE vs Contrastive) and total number of mini-batch augmentations ($K=32$ vs $K=2$). In both conditions we observe a severe degradation of the performance with respect to the complete model (from a minimum of $-3\%$ to a maximum of $-23\%$), confirming that the relation module is a fundamental component in the pipeline (see discussion in Section~\ref{sec:discussion}).

\textbf{Qualitative analysis.} In Appendix~\ref{appendix:additional_retrieval} is presented a qualitative comparison between the proposed method and RotationNet, on an image retrieval downstream task. Given a random query image (not cherry-picked) the top-10 most similar images in representation space are retrieved. Our method shows better distinction between categories which are hard to separate (e.g. ships vs planes, trucks vs cars). The lower sample variance and the higher similarity with the query, confirm the fine-grained organization of the representations, which account for color, texture, and geometry. An analysis of retrieval errors in Appendix~\ref{appendix:additional_retrieval_error} shows that the proposed method is superior in accuracy across all categories while being more robust against misclassification, with a top-10 retrieval accuracy of 67.8\% against 47.7\% of RotationNet. In Appendix~\ref{appendix:additional_qualitative} we report a qualitative analysis of the representations (ResNet-32, CIFAR-10) using t-SNE \citep{maaten2008visualizing}. Relational reasoning is able to aggregate the data in a more effective way, and to better capture high level relations with lower scattering (e.g. vehicles vs animals super-categories).

\section{Discussion and conclusions}\label{sec:discussion}

Self-supervised relational reasoning is effective on a wide range of tasks in both a quantitative and qualitative manner, and with backbones of different size (ResNet-32, ResNet-56 and ResNet-34, with $0.5 \times 10^{6}$, $0.9 \times 10^{6}$ and $21.3 \times 10^{6}$ parameters). 
Representations learned through comparison can be easily transferred across domains, they are fine-grained and compact, which may be due to the direct correlation between accuracy and number of augmentations. An instance is pushed towards a positive neighborhood (intra-reasoning) and repelled from a complementary set of negatives (inter-reasoning). The number of augmentations may have a primary role in this process affecting the quality of the clusters.
The possibility to exploit an high number of augmentations, by generating them on the fly, could be decisive in the low-data regime (e.g. unsupervised few-shot/online learning) where self-supervised relational reasoning has the potential to thrive. Those are factors that require further consideration and investigation.

\textbf{From self-supervised to supervised.} Recent work has showed that contrastive learning can be used in a supervised setting with competitive results \citep{prannay2020supervised}. In our experiments we have observed a similar trend, with relational reasoning approaching the supervised performance when all the labels are available. However, we have obtained those results using the same hyperparameters and augmentations used in the self-supervised case, while there may be alternatives that are more effective. Learning by comparison could help in disentangling fine-grained differences in a fully supervised setting with high number of classes, and be decisive to build complex taxonomic representations, as pointed out in cognitive studies \citep{gentner1999comparison, namy2002making}.

\textbf{Comparison with contrastive methods.} We have compared relational reasoning to a state-of-the-art contrastive learning method (SimCLR) using the same backbone, head, augmentation strategy, and learning schedule. Relational reasoning outperforms SimCLR (+3\% on average) using a lower number of pairs, being more efficient. Given a mini-batch of size 64, relational reasoning uses $6.35 \times 10^{4}$ ($K=32$) and $1.5 \times 10^{4}$ ($K=16$) pairs, against $6.55\times10^{4}$ of SimCLR with mini-batch 128.
Contrastive losses needs a large number of negatives, which can be gathered by increasing $M$ the size of the mini-batch, or increasing $K$ the number of augmentations (both solutions incur a quadratic cost, see Section~\ref{ssec:inputs_augmentation}). High quality negatives can only be gathered following the first solution, since the second provides lower sample variance. A typical mini-batch in SimCLR encloses $98\%$ negatives and $2\%$ positives, in our method $50\%$ negatives and $50\%$ positives. The larger set of positives could be one of the reasons why relational reasoning is more effective in disentangling fine-grained representations.
In addition to the difference in loss type, there is an important structural difference between the two approaches: in SimCLR pairs are allocated in the loss space and then compared via dot product, while in relational reasoning they are aggregated in the space of transferable representations and compared through a relation head. Ablation studies in Section~\ref{sec:experiments} have shown that this structural difference is fundamental for obtaining higher performances, but the way it influences the learning dynamics and the optimization process is not clear and requires further investigation.

\textbf{Why does cross-entropy work so well?}
We argue that in the context of recent state-of-the-art methods, cross-entropy has been overlooked in favor of contrastive losses. Our experiments show that cross-entropy is a more efficient and effective objective function with respect to the commonly used contrastive losses. Based on the results of the ablation studies, we hypothesize that the difference in performance is mainly due to the use of a relation module in conjunction with the binary cross-entropy loss. When the BCE is split from the relation head and applied directly to the representations there is a drastic drop in performance; applying the BCE to surrogate representations in a second encoding stage (like in SimCLR) is equally ineffective. Therefore, the use of BCE on its own does not provide any advantage but in concert with the relation head it becomes effective. A more thorough analysis is necessary to substantiate these findings, which is left for future work.

\section*{Broader Impact}

The motivation behind this work is to build systems able to exploit a large amount of unlabeled data. Applications that could benefit from the proposed method span from standard supervised classifiers to medical diagnostic systems. Therefore, there is a large number of individuals who may benefit or be harmed from this research. This requires putting some effort into selecting the data source, especially when the system is scaled. 

In most cases a large body of unlabeled images can be easily gathered from the internet; to avoid biases those images should be representative of different categories. Our method does not guarantee unbiased predictions, therefore it should be used with caution in critical applications. Individuals who may want to use it should consider the particular source of data at hand and evaluate how it could impact the system performance after the final deployment.

\begin{ack}
This work was supported by a Huawei DDMPLab Innovation Research Grant.

MP and AS would like to thank anonymous reviewers for useful comments and suggestions; the BayesWatch team for feedback and discussion, in particular Elliot J. Crowley, Luke Darlow, and Joseph Mellor. MP would like to thank the Becchi team for revising the preliminary version of the manuscript, in particular Valerio Biscione, Riccardo Polvara, and Luca Surace.

\end{ack}

\small
\bibliographystyle{apalike}
\bibliography{references.bib}
\FloatBarrier
\clearpage
\appendix 

\section{Implementation details}\label{appendix:implementation_details}
\FloatBarrier

\subsection{Datasets}

For datasets with low/medium number of categories we used CIFAR-10 and CIFAR-100~\citep{krizhevsky2009learning}, which are composed of $32 \times 32$ RGB images, with 10 and 100 classes respectively. In addition we used the 20 super-classes of CIFAR-100 (naming this CIFAR-100-20), which consists of broader categories (e.g. fruits, mammals, insects, etc). In the finetuning experiments we used the STL-10 dataset \citep{coates2011analysis} which provides 100K RGB images of size $96 \times 96$ in the unlabeled set, 5K images in the labeled set, and 8K images in the test set.

For datasets with an high number of categories we used the tiny-ImageNet and SlimageNet~\citep{antoniou2020defining} datasets, both of them derived from ImageNet~\citep{russakovsky2015imagenet}. Tiny-ImageNet consists of 200 different categories, with 500 training images ($64 \times 64$, 100K in total), 50 validation images (10K in total), and 50 test images (10K in total). SlimageNet consists of $64 \times 64$ RGB images, 1000 categories with 160 training images (160K in total), 20 validation images (20K in total), and 20 test images (20K in total). Both of them are considered more challenging than ImageNet because of the lower resolution of the images and lower number of training samples.

\subsection{Backbones}

We use off-the-shelf Pytorch implementations of ResNets as described in the original paper \citep{he2016deep}. Some of these networks have quite different structure, with ResNet-8/32/56 based on three hyper-blocks (ResNet-32 has $0.5 \times 10^{6}$ total parameters) and ResNet-34 based on four hyper-blocks ($21.3 \times 10^{6}$ total parameters).
The Conv-4 backbone is based on three blocks (8, 16, 32 feature maps), each one performing: convolution (kerne-size=3, stride=1, padding=1), BatchNorm, ReLU, average pooling (kerne-size=2, stride=2). The fourth block (64 feature maps) performed the same operations but with an adaptive average pooling to squeeze the maps to unit shape in the spatial dimension.
We used standard fan-in/fan-out weight initialization, and set BatchNorm weights to 1 and bias to 0. For Conv-4 and ResNet-8/32/56 the size of the representations is 64, whereas for ResNet-34 is 512.

\subsection{Augmentations}\label{appendix:augmentations}

During the self-supervised training phase of our method we used a set of augmentations which is similar to the one adopted by \cite{chen2020simple}. We apply horizontal flip (50\% chance), random crop-resize, conversion to grayscale (20\% chance), and color jitter (80\% chance). Random crop-resize consists of cropping the given image (from 0.08 to 1.0 of the original size), changing the aspect ratio (from 3/4 to 4/3 of the original aspect ratio), and finally resizing to input shape using a bilinear interpolation. Color jitter consists of sampling from a uniform distribution $[0, max]$ a jittering value for: brightness ($max=0.8$), contrast ($max=0.8$), saturation ($max=0.8$), and hue ($max=0.2$).

\subsection{Computing infrastructure} \label{appendix:computing_infrastracture}

All the experiments have been performed on a workstation with 20 cores, 187 GB of RAM, and with 8 NVIDIA GeForce RTX 2080 Ti GPUs (11 GB of internal RAM). All the methods could fit on a single one of those GPUs.

\subsection{Other methods}

\textbf{Supervised.} This baseline consists of standard supervised training. We used standard data augmentation (horizontal flip and random crop) and learning schedule (SGD optimizer with initial learning rate of 0.1 divided by 10 at 50\% and 75\% of total epochs). It represents an upper bound. When evaluated for the number of augmentations (Appendix~\ref{appendix:number_augmentations}) the same strategy adopted in our method (Appendix~\ref{appendix:augmentations}) has been used to augment the input mini-batch (size 128) $K$ times with coherent labels.

\textbf{Random weights.} This baseline consists of initializing the weights of the backbone via standard fan-in/fan-out, then perform linear evaluation optimizing the last linear layer (without backpropagation on the backbone). It represents a lower bound since the backbone is not trained.

\textbf{DeepCluster \citep{caron2018deep}.}  We adapted the open-source implementation provided by the authors\footnote{\url{https://github.com/facebookresearch/deepcluster}}. Clustering has been performed at the beginning of each epoch by using the k-means algorithm available in Scikit-learn. We performed whitening over the features before the clustering step, as suggested by the authors. We used a number of cluster one order of magnitude larger than the number of classes in the dataset, as recommended by the authors to improve the performance. We also used an MLP head (256 hidden units with leaky-ReLU and BatchNorm) instead of a linear layer, since in our tests this showed to slightly boost the performance. The MLP weights have been reset at the beginning of each epoch as in the original code. We optimized the model minimizing the cross-entropy loss between the pseudo-labels provided by the clustering and the network outputs. We used Adam optimizer with learning rate $10^{-3}$.

\textbf{RotationNet \citep{gidaris2018unsupervised}.} Given the simplicity of the method, this has been reproduced locally following the instructions of the authors. Labels are provided by 4 rotations ($0^{\circ}, 90^{\circ}, 180^{\circ}, 270^{\circ}$), those are the one providing the highest accuracy according to the authors. The input mini-batch of size 128, has been augmented adding 4 rotations for each image (resulting in a mini-batch of size $128 \times 4$). This is in line with the best performing strategy reported by the authors. In all experiments the cross-entropy loss between the network output and the labels provided by the rotation has been minimized (Adam optimizer, learning rate $10^{-3}$). When evaluated for the number of augmentations (Appendix~\ref{appendix:number_augmentations}) the same strategy used in our method has been applied (Appendix~\ref{appendix:augmentations}), augmenting the input mini-batch (size 128) $K$ times with coherent self-supervised rotation labels. In order to keep the size of the mini-batch manageable the additional 4 rotations for image have not been included, since this would increase the size to $4 \times K$ and not fit on the available hardware.

\textbf{Deep InfoMax \citep{hjelm2018learning}} The code has been adapted from open-source implementations available online (see code for details) and from the code provided by the authors\footnote{\url{https://github.com/rdevon/DIM}}. The local version of the algorithm has been used ($\alpha=0, \beta=1.0, \gamma=0.1$), as reported by the authors this is the one with the best performance. The capacity of the discriminator networks has been partially reduced to fit the available hardware and to speedup the training, this did not affected significantly the results. We used Adam optimizer with learning rate $10^{-4}$ as in the original paper.

\textbf{SimCLR \citep{chen2020simple}} The code has been adapted from the implementation provided by the authors\footnote{\url{https://github.com/google-research/simclr}} and other open-source implementations (see code for details). To have a fair comparison with our method we used the same MLP head, the same data augmentation strategy, and optimizer (Adam with learning rate$10^{-3}$). We used a temperature of 0.5 in all the experiments, this was reported as the consistent optimal value regardless of the batch sizes in the original paper. We could not replicate the original setup reported by the authors on very large mini-batches, since it is computationally expensive, requiring 32 to 128 cores on Tensor Processing Units (TPUs). We adapted the setup to our available hardware (described in Appendix~\ref{appendix:computing_infrastracture}), and we guaranteed a fair comparison by using a comparable number of pairs. In particular, we used a mini-batch of 128 images, which results in $6.5 \times 10^{4}$ pairs, this is similar (or superior) to the number of pairs compared by our method which is $6.3 \times 10^{4}$ for $K=32$, $3.8\times10^{4}$ for $K=25$, and $1.5\times10^{4}$ for $K=16$.

\subsection{Finetuning experiments}\label{appendix:implementation_finetuning}

All methods are trained for 300 epochs on the unlabeled portion of the STL-10 dataset, using the same hyperparameters and augmentations described before and a ResNet34 backbone. In the finetuning stage the pretrained backbone is coupled with a linear classifier and both of them are trained using Adam optimizer for 100 epochs with mini-batch of size 32. A lower learning rate for the backbone ($10^{-4}$) respect to the linear classifier ($10^{-3}$) has been used. Both learning rates are divided by 10 at 50\% and 75\% of total epochs. The same augmentations of \cite{ji2019invariant} have been used for the finetuning stage. Those consists of affine transformations (50\% chance) sampled from a uniform distribution $[min, max]$: random rotation ($min=-18^{\circ}, max=18^{\circ}$), scale ($min=0.9, max=1.1$), translation ($min=0, max=0.1$), shear ($min=-10, max=10$), and bilinear interpolation, cutout (50\% chance) with $32\times32$ patches. Same schedule and augmentations have also been used to train (from scratch) the supervised baseline on the labeled set of data (100 epochs).

\subsection{Semi-supervised experiments}

We adapted our method to the semi-supervised case by coupling instances sampled from the same category. Those instances represented a portion of the total number of pairs in the mini-batch depending on the percentage of available labels.
Results for each conditions are the average of three seeds. We used the same hyperparameters described in the linear evaluation phase.
We did not prevent possible collisions between classes during the allocation of negative pairs. Collisions are unlikely in datasets with medium/high number of classes, but a slight performance improvement could be obtained if negatives are paired without collisions. 

\subsection{Qualitative analysis experiments}

For the qualitative analysis we compared the representations generated by the supervised baseline, Rotation Net, and our method on CIFAR-10 with a ResNet-32 backbone at the end of the training (200 epochs). The query images were randomly sampled and the representations compared using Euclidean distance.
For the t-SNE analysis we used the Scikit implementation of the algorithms and set the hyperparameters as follows: 1000 iterations, Euclidean metric, random init, perplexity 30, learning rate 200.

\section{Additional results}\label{appendix:additional_results}
\FloatBarrier

\subsection{Linear evaluation}\label{appendix:additional_linear_evaluation}

\begin{table}[H]
 \caption{Linear evaluation. Self-supervised training on unlabeled data and linear evaluation on labeled data. Comparison between three datasets (CIFAR-10, CIFAR-100, tiny-ImageNet) for a shallow (Conv-4) and a deep (ResNet-32) backbone. Mean accuracy (percentage) and standard deviation over three runs. Best results highlighted in bold.}
 \label{tab:linear_evaluation}
 \begin{adjustbox}{width=\columnwidth,center}
  \centering
  \begin{tabular}{lcccccc}
    \toprule
     & \multicolumn{3}{c}{\textbf{Conv-4}} & \multicolumn{3}{c}{\textbf{ResNet-32}} \\
    \cmidrule[0.1pt](r){2-4} \cmidrule[0.1pt](l){5-7}
    \textbf{Method} &
    \textbf{CIFAR-10} & \textbf{CIFAR-100} & \textbf{tiny-ImageNet} &
    \textbf{CIFAR-10} & \textbf{CIFAR-100} & \textbf{tiny-ImageNet} \\
    \midrule
    Supervised (upper bound) & 
    80.46$\pm$\small{0.39} & 49.29$\pm$\small{0.85}  & 36.47$\pm$\small{0.36} &
    90.87$\pm$\small{0.41} & 65.32$\pm$\small{0.22}  & 50.09$\pm$\small{0.32} \\
    Random Weights (lower bound) & 
    32.92$\pm$\small{1.88} & 10.79$\pm$\small{0.59}  & 6.19$\pm$\small{0.13} &
    27.47$\pm$\small{0.83} &  7.65$\pm$\small{0.44}  & 3.24$\pm$\small{0.43} \\
    \cmidrule(l){1-7}
    DeepCluster \citep{caron2018deep} & 
    42.88$\pm$\small{0.21} & 21.03$\pm$\small{1.56}  & 12.60$\pm$\small{1.23} &
    43.31$\pm$\small{0.62} & 20.44$\pm$\small{0.80}  & 11.64$\pm$\small{0.21} \\
    RotationNet \citep{gidaris2018unsupervised} & 
    56.73$\pm$\small{1.71} & 27.45$\pm$\small{0.80}  & 18.40$\pm$\small{0.95} &
    62.00$\pm$\small{0.79} & 29.02$\pm$\small{0.18}  & 14.73$\pm$\small{0.48} \\
    Deep InfoMax \citep{hjelm2018learning} & 
    44.60$\pm$\small{0.27} & 22.74$\pm$\small{0.21}  & 14.19$\pm$\small{0.13} &
    47.13$\pm$\small{0.45} & 24.07$\pm$\small{0.05}  & 17.51$\pm$\small{0.15} \\
    SimCLR \citep{chen2020simple} & 
    60.43$\pm$\small{0.26} & 30.45$\pm$\small{0.41}  & 20.90$\pm$\small{0.15} &
    \textbf{77.02$\pm$\small{0.64}} & 42.13$\pm$\small{0.35}  & 25.79$\pm$\small{0.40} \\  
    \emph{Relational Reasoning} (ours)  & \textbf{61.03$\pm$\small{0.23}} & \textbf{33.38$\pm$\small{1.02}}  & \textbf{22.31$\pm$\small{0.19}} &
    74.99$\pm$\small{0.07} & \textbf{46.17$\pm$\small{0.16}}  & \textbf{30.54$\pm$\small{0.42}} \\
    \bottomrule
  \end{tabular}
 \end{adjustbox}
\end{table}

\begin{table}[H]
 \caption{Linear evaluation on SlimageNet~\citep{antoniou2020defining}. This dataset is more challenging than ImageNet, since it only has 160 low-resolution ($64 \times 64$) color images for each one of the 1000 classes of ImageNet. Below is reported the linear evaluation accuracy on labeled data with a ResNet-32 backbone, after training on unlabeled data. Mean accuracy (percentage) and standard deviation over three runs. Best result highlighted in bold.}
 \label{tab:linear_evaluation_slim}
 \begin{adjustbox}{width=0.40\columnwidth,center}
  \centering
  \begin{tabular}{lc}
    \toprule
    \textbf{Method} & \textbf{SlimageNet} \\
    \midrule
    Supervised (upper bound) & 33.94$\pm$\small{0.21} \\
    Random Weights (lower bound) & 0.79$\pm$\small{0.09} \\
    \midrule
    RotationNet \citep{gidaris2018unsupervised} & 7.25$\pm$\small{0.28} \\
    SimCLR \citep{chen2020simple} & 14.32$\pm$\small{0.24} \\
    \emph{Relational Reasoning} (ours)  & \textbf{15.81$\pm$\small{0.72}} \\
    \bottomrule
  \end{tabular}
 \end{adjustbox}
\end{table}

\subsection{Domain transfer}\label{appendix:additional_domain_transfer}

\begin{table}[H]
 \caption{Domain transfer. Training with self-supervision on unlabeled CIFAR-10 linear evaluation on CIFAR-100 (10 $\rightarrow$ 100), and viceversa (100 $\rightarrow$ 10). $\Delta$ indicates the difference between the accuracy in the standard setting (unsupervised train and linear evaluation on the same dataset) and the accuracy in the transfer setting (unsupervised train on first dataset and linear evaluation on the second dataset). Mean accuracy (percentage) and standard deviation over three runs. Best results highlighted in bold.}
 \label{tab:transfer_learning}
 \begin{adjustbox}{width=\columnwidth,center}
  \centering
  \begin{tabular}{lcccccccc}
    \toprule
    %\cmidrule[1pt](r){2-7}
     & \multicolumn{4}{c}{\textbf{Conv-4}} & \multicolumn{4}{c}{\textbf{ResNet-32}} \\
    \cmidrule[0.1pt](r){2-5} \cmidrule[0.1pt](l){6-9}
    \textbf{Method} &
    \textbf{10 $\rightarrow$ 100} & $\boldsymbol{\Delta}$ & \textbf{100 $\rightarrow$ 10} & $\boldsymbol{\Delta}$ &
    \textbf{10 $\rightarrow$ 100} & $\boldsymbol{\Delta}$ & \textbf{100 $\rightarrow$ 10} & $\boldsymbol{\Delta}$\\
    \midrule
    %\cmidrule(r){1-5}
    Supervised (upper bound) & 
    32.06$\pm$\small{0.63} & -17.23 & 64.00$\pm$\small{1.07} & -16.46 &
    33.98$\pm$\small{0.70} & -31.34 & 71.01$\pm$\small{0.44} & -19.86 \\
    Random Weights (lower bound) & 
    10.79$\pm$\small{0.59} & n/a & 32.92$\pm$\small{1.89} & n/a &
    7.65$\pm$\small{0.44} & n/a & 27.47$\pm$\small{0.83} & n/a \\
    \cmidrule(l){1-9}
    DeepCluster \citep{caron2018deep} & 
    19.68$\pm$\small{1.23} & -1.35 & 43.59$\pm$\small{1.31} & +0.71 &
    18.37$\pm$\small{0.41} & -2.07 & 43.39$\pm$\small{1.84} & +0.08 \\
    RotationNet \citep{gidaris2018unsupervised} & 
    26.06$\pm$\small{0.09} & -1.39 & 51.86$\pm$\small{0.36} & -4.87 &
    27.02$\pm$\small{0.20} & -2.00 & 52.22$\pm$\small{0.70} & -9.78 \\
    Deep InfoMax \citep{hjelm2018learning} & 
    22.35$\pm$\small{0.12} & -0.39 & 43.30$\pm$\small{0.15} & -1.30 &
    23.73$\pm$\small{0.04} & -0.34 & 45.05$\pm$\small{0.24} & -2.08 \\
    SimCLR \citep{chen2020simple} &
    29.20$\pm$\small{0.08} & -1.25 & 54.73$\pm$\small{0.60} & -5.70 &
    36.21$\pm$\small{0.16} & -5.92 & 65.59$\pm$\small{0.76} & -11.43 \\
    \emph{Relational Reasoning} (ours) & 
    \textbf{31.84$\pm$\small{0.23}} & -1.54 & \textbf{57.30$\pm$\small{0.26}} & -3.73 &
    \textbf{41.50$\pm$\small{0.35}} & -4.67 & \textbf{67.81$\pm$\small{0.42}} & -7.18 \\
    \bottomrule
  \end{tabular}
 \end{adjustbox}
\end{table}

\subsection{Grain}\label{appendix:additional_grain}

\begin{table}[H]
 \caption{Grain. Training with self-supervision on unlabeled CIFAR-100, and linear evaluation on labeled CIFAR-100 Fine-Grained (100 classes) and CIFAR-100-20 Coarse-Grained (20 super-classes). Mean accuracy (percentage) and standard deviation over three runs. Best results highlighted in bold.}
 \label{tab:fine_coarse}
 \begin{adjustbox}{width=0.85\columnwidth,center}
  \centering
  \begin{tabular}{lcccc}
    \toprule
     & \multicolumn{2}{c}{\textbf{Conv-4}} & \multicolumn{2}{c}{\textbf{ResNet-32}} \\
    \cmidrule[0.1pt](r){2-3} \cmidrule[0.1pt](l){4-5}
    \textbf{Method} &
    \textbf{Fine-Grain} & \textbf{Coarse-Grain} &
    \textbf{Fine-Grain} & \textbf{Coarse-Grain} \\
    \midrule
    Supervised (upper bound) & 
    49.29$\pm$\small{0.85}  & 59.91$\pm$\small{0.62} &
    65.32$\pm$\small{0.22}  & 76.35$\pm$\small{0.57} \\
    Random Weights (lower bound) & 
    10.79$\pm$\small{0.59} & 19.94$\pm$\small{0.31} &
    7.65$\pm$\small{0.44}  & 16.56$\pm$\small{0.48} \\
    \cmidrule(l){1-5}
    DeepCluster \citep{caron2018deep} & 
    21.03$\pm$\small{1.56}  & 30.07$\pm$\small{2.06} &
    20.44$\pm$\small{0.80}  & 29.49$\pm$\small{1.36} \\
    RotationNet \citep{gidaris2018unsupervised} & 
    27.45$\pm$\small{0.80}  & 35.49$\pm$\small{0.17} &
    29.02$\pm$\small{0.19}  & 40.45$\pm$\small{0.39} \\
    Deep InfoMax \citep{hjelm2018learning} & 
    22.74$\pm$\small{0.21}  & 32.36$\pm$\small{0.43} &
    24.07$\pm$\small{0.05}  & 33.92$\pm$\small{0.34} \\
    SimCLR \citep{chen2020simple} & 
    30.45$\pm$\small{0.41}  & 37.72$\pm$\small{0.14} &
    42.13$\pm$\small{0.35}  & 51.88$\pm$\small{0.48} \\    
    \emph{Relational Reasoning} (ours) & 
    \textbf{33.38$\pm$\small{1.02}}  & \textbf{40.86$\pm$\small{1.03}} &
    \textbf{46.17$\pm$\small{0.17}}  & \textbf{52.44$\pm$\small{0.47}} \\
    \bottomrule
  \end{tabular}
 \end{adjustbox}
\end{table}

\subsection{Finetuning}\label{appendix:additional_finetuning}

\begin{table}[H]
 \caption{Finetuning. Comparison with other results reported in the literature on unsupervised training and finetuning on the STL-10 dataset. Best result in bold. \emph{Local} refers to our local reproduction of the method, with results reported as \emph{best (mean $\pm$ std)} on three runs with different seeds. Note that backbone and learning schedule may differ. The ResNet-34 backbone is much larger than ResNet-32 ($21.3 \times 10^{6}$ vs $0.47 \times 10^{6}$), showing that the proposed method can be effectively scaled.}
 \label{tab:additional_finetuning}
 \begin{adjustbox}{width=1.0\columnwidth,center}
  \centering
  \begin{tabular}{lccc}
    \toprule
    \cmidrule[0.1pt](r){1-4}
    \textbf{Method} & \textbf{Reference} &
    \textbf{Backbone} & \textbf{Accuracy} \\
    \midrule
    Supervised (crop + cutout) & \cite{devries2017improved} &
    WideResnet-16-8  & 87.30\\
    Supervised (scattering) & \cite{oyallon2017scaling} &
    Hybrid-WideResnet  & 87.60\\
    Exemplars \citep{dosovitskiy2014discriminative} & \cite{dosovitskiy2014discriminative} & Conv-3 & 72.80\\
    Artifacts \citep{jenni2018self} & \cite{jenni2018self} & Custom & 80.10\\
    ADC \citep{haeusser2018associative} & \cite{ji2019invariant} & ResNet-34 & 56.70 \\
    DeepCluster \citep{caron2018deep} & \cite{ji2019invariant} &
    ResNet-34  & 73.40\\
    Deep InfoMax \citep{hjelm2018learning} & \cite{ji2019invariant} &
    AlexNet  & 77.00\\
    Invariant Info Clustering \citep{ji2019invariant} & \cite{ji2019invariant} & ResNet-34 & 88.80\\
    %\cmidrule(l){1-4}    
    Supervised (affine + cutout) & Local &
    ResNet-34  & 72.04 (69.82 $\pm$ \small{3.36})\\
    DeepCluster \citep{caron2018deep} & Local &
    ResNet-34  & 74.00 (73.37 $\pm$ \small{0.55})\\
    RotationNet \citep{gidaris2018unsupervised} & Local &
    ResNet-34  & 83.77 (83.29 $\pm$ \small{0.44})\\
    Deep InfoMax \citep{hjelm2018learning} & Local &
    ResNet-34  & 76.45 (76.03 $\pm$ \small{0.37})\\
    SimCLR \citep{chen2020simple} & Local &
    ResNet-34  & 89.44 (89.31 $\pm$ \small{0.14})\\    
    \emph{Relational Reasoning} (ours) & Local &
    ResNet-34  & \textbf{90.04} (\textbf{89.67 $\pm$ \small{0.33}})\\
    \bottomrule
  \end{tabular}
 \end{adjustbox}
\end{table}

\subsection{Performance with different backbones}\label{appendix:additional_backbones}

\begin{table}[H]
 \caption{Comparison on different backbones: linear evaluation. Comparison between four backbones of different depth for baselines and the three best performing methods. Training with self-supervision on unlabeled CIFAR-10 and CIFAR-100, and linear evaluation on labeled version of the same datasets. Mean accuracy (percentage) and standard deviation over three runs. Best results highlighted in bold.}
 \label{tab:backbones_linear}
 \begin{adjustbox}{width=1.0\columnwidth,center}
  \centering
  \begin{tabular}{lcccccccc}
    \toprule
     & \multicolumn{2}{c}{\textbf{Conv-4}} & \multicolumn{2}{c}{\textbf{ResNet-8}} & \multicolumn{2}{c}{\textbf{ResNet-32}} & \multicolumn{2}{c}{\textbf{ResNet-56}}\\
    \cmidrule[0.1pt](r){2-3} \cmidrule[0.1pt](l){4-5} \cmidrule[0.1pt](l){6-7} \cmidrule[0.1pt](l){8-9}
    \textbf{Method} &
    \textbf{CIFAR-10} & \textbf{CIFAR-100} &
    \textbf{CIFAR-10} & \textbf{CIFAR-100} &
    \textbf{CIFAR-10} & \textbf{CIFAR-100} &
    \textbf{CIFAR-10} & \textbf{CIFAR-100} \\
    \midrule
    Supervised (upper bound) & 
    80.46$\pm$\small{0.39} & 49.29$\pm$\small{0.85} &
    87.08$\pm$\small{0.17} & 59.41$\pm$\small{1.15} &
    90.87$\pm$\small{0.41} & 65.32$\pm$\small{0.22} &
    91.40$\pm$\small{0.30} & 67.54$\pm$\small{0.32} \\
    Random Weights (lower bound) & 
    32.92$\pm$\small{1.89} & 10.79$\pm$\small{0.59} &
    35.94$\pm$\small{1.39} & 13.08$\pm$\small{0.91} &
    27.47$\pm$\small{0.83} &  7.65$\pm$\small{0.44} &
    13.53$\pm$\small{3.66} &  1.88$\pm$\small{0.14} \\
    \cmidrule(l){1-9}
    RotationNet \citep{gidaris2018unsupervised} & 
    56.73$\pm$\small{1.71} & 27.45$\pm$\small{0.80} &
    62.73$\pm$\small{0.94} & 32.09$\pm$\small{0.87} &
    62.00$\pm$\small{0.79} & 29.02$\pm$\small{0.18} &
    61.66$\pm$\small{1.11} & 28.24$\pm$\small{0.23} \\
    SimCLR \citep{chen2020simple} & 
    60.43$\pm$\small{0.26} & 30.45$\pm$\small{0.41} &
    \textbf{69.85$\pm$\small{0.58}} & 36.23$\pm$\small{0.15} &
    \textbf{77.02$\pm$\small{0.64}} & 42.13$\pm$\small{0.35} &
    \textbf{78.75$\pm$\small{0.24}} & 44.33$\pm$\small{0.48} \\    
    \emph{Relational Reasoning} (ours) & 
    \textbf{61.03$\pm$\small{0.23}} & \textbf{33.38$\pm$\small{1.02}} &
    67.97$\pm$\small{0.58} & \textbf{38.18$\pm$\small{0.63}} &
    74.99$\pm$\small{0.07} & \textbf{46.17$\pm$\small{0.17}} &
    77.51$\pm$\small{0.00} & \textbf{47.90$\pm$\small{0.27}} \\
    \bottomrule
  \end{tabular}
 \end{adjustbox}
\end{table}

\begin{table}[H]
 \caption{Comparison on different backbones: grain. Comparison between four backbones of different depth for baselines and the three best performing methods. Training with self-supervision on unlabeled CIFAR-100 and linear evaluation on labeled version of the same datasets with 100 labels (fine) or 20 super-labels (coarse). Mean accuracy (percentage) and standard deviation over three runs. Best results highlighted in bold.}
 \label{tab:backbones_grain}
 \begin{adjustbox}{width=1.0\columnwidth,center}
  \centering
  \begin{tabular}{lcccccccc}
    \toprule
     & \multicolumn{2}{c}{\textbf{Conv-4}} & \multicolumn{2}{c}{\textbf{ResNet-8}} & \multicolumn{2}{c}{\textbf{ResNet-32}} & \multicolumn{2}{c}{\textbf{ResNet-56}}\\
    \cmidrule[0.1pt](r){2-3} \cmidrule[0.1pt](l){4-5} \cmidrule[0.1pt](l){6-7} \cmidrule[0.1pt](l){8-9}
    \textbf{Method} &
    \textbf{Fine} & \textbf{Coarse} &
    \textbf{Fine} & \textbf{Coarse} &
    \textbf{Fine} & \textbf{Coarse} &
    \textbf{Fine} & \textbf{Coarse} \\
    \midrule
    Supervised (upper bound) & 
    49.29$\pm$\small{0.85} & 59.91$\pm$\small{0.62} & 
    59.41$\pm$\small{1.15} & 70.12$\pm$\small{0.33} & 
    65.32$\pm$\small{0.22} & 76.35$\pm$\small{0.57} & 
    67.54$\pm$\small{0.32} & 77.60$\pm$\small{0.43} \\
    Random Weights (lower bound) & 
    10.79$\pm$\small{0.59} & 19.94$\pm$\small{0.31} & 
    13.08$\pm$\small{0.91} & 23.12$\pm$\small{0.90} & 
     7.65$\pm$\small{0.44} & 16.56$\pm$\small{0.48} &  
     1.88$\pm$\small{0.14} &  6.88$\pm$\small{0.35} \\
    \cmidrule(l){1-9}
    RotationNet \citep{gidaris2018unsupervised} & 
    27.45$\pm$\small{0.80} & 35.49$\pm$\small{0.17} & 
    32.09$\pm$\small{0.87} & 41.21$\pm$\small{0.94} & 
    29.02$\pm$\small{0.18} & 40.45$\pm$\small{0.39} & 
    28.24$\pm$\small{0.23} & 39.16$\pm$\small{0.35} \\
    SimCLR \citep{chen2020simple} & 
    30.45$\pm$\small{0.41} & 37.72$\pm$\small{0.14} & 
    36.23$\pm$\small{0.15} & 43.78$\pm$\small{0.92} & 
    42.13$\pm$\small{0.35} & 51.87$\pm$\small{0.48} & 
    44.33$\pm$\small{0.48} & 54.09$\pm$\small{0.15} \\    
    \emph{Relational Reasoning} (ours) & 
    \textbf{33.38$\pm$\small{1.02}} & \textbf{40.86$\pm$\small{1.03}} & 
    \textbf{38.18$\pm$\small{0.63}} & \textbf{45.36$\pm$\small{0.55}} & 
    \textbf{46.17$\pm$\small{0.17}} & \textbf{52.44$\pm$\small{0.47}} & 
    \textbf{47.90$\pm$\small{0.27}} & \textbf{54.90$\pm$\small{0.07}} \\
    \bottomrule
  \end{tabular}
 \end{adjustbox}
\end{table}

\begin{table}[H]
 \caption{Comparison on different backbones: domain transfer. Comparison between four backbones of different depth for baselines and the three best performing methods. Training with self-supervision on unlabeled CIFAR-10 linear evaluation on CIFAR-100 (10 $\rightarrow$ 100), and viceversa (100 $\rightarrow$ 10). Mean accuracy (percentage) and standard deviation over three runs. Best results highlighted in bold.}
 \label{tab:backbones_domain_transfer}
 \begin{adjustbox}{width=1.0\columnwidth,center}
  \centering
  \begin{tabular}{lcccccccc}
    \toprule
     & \multicolumn{2}{c}{\textbf{Conv-4}} & \multicolumn{2}{c}{\textbf{ResNet-8}} & \multicolumn{2}{c}{\textbf{ResNet-32}} & \multicolumn{2}{c}{\textbf{ResNet-56}}\\
    \cmidrule[0.1pt](r){2-3} \cmidrule[0.1pt](l){4-5} \cmidrule[0.1pt](l){6-7} \cmidrule[0.1pt](l){8-9}
    \textbf{Method} &
    \textbf{10$\rightarrow$100} & \textbf{100$\rightarrow$10} &
    \textbf{10$\rightarrow$100} & \textbf{100$\rightarrow$10} &
    \textbf{10$\rightarrow$100} & \textbf{100$\rightarrow$10} &
    \textbf{10$\rightarrow$100} & \textbf{100$\rightarrow$10} \\
    \midrule
    Supervised (upper bound) & 
    32.06$\pm$\small{0.63} & 64.00$\pm$\small{1.07} & 
    36.83$\pm$\small{0.36} & 71.20$\pm$\small{0.18} & 
    33.98$\pm$\small{0.70} & 71.01$\pm$\small{0.44} & 
    33.92$\pm$\small{0.50} & 71.97$\pm$\small{0.17} \\
    Random Weights (lower bound) & 
    10.79$\pm$\small{0.59} & 32.92$\pm$\small{1.89} & 
    13.08$\pm$\small{0.91} & 35.94$\pm$\small{1.39} & 
     7.65$\pm$\small{0.44} & 27.47$\pm$\small{0.83} &  
     1.88$\pm$\small{0.14} & 13.53$\pm$\small{3.66} \\
    \cmidrule(l){1-9}
    RotationNet \citep{gidaris2018unsupervised} & 
    26.06$\pm$\small{0.09} & 51.86$\pm$\small{0.36} & 
    31.60$\pm$\small{0.54} & 56.85$\pm$\small{0.13} & 
    27.02$\pm$\small{0.20} & 52.22$\pm$\small{0.70} & 
    27.25$\pm$\small{0.62} & 51.82$\pm$\small{0.58} \\
    SimCLR \citep{chen2020simple} & 
    29.20$\pm$\small{0.08} & 54.73$\pm$\small{0.60} & 
    34.46$\pm$\small{0.78} & 61.34$\pm$\small{0.24} & 
    36.21$\pm$\small{0.16} & 65.59$\pm$\small{0.76} & 
    36.79$\pm$\small{0.45} & 66.19$\pm$\small{0.80} \\    
    \emph{Relational Reasoning} (ours) & 
    \textbf{31.84$\pm$\small{0.23}} & \textbf{57.30$\pm$\small{0.26}} & 
    \textbf{36.07$\pm$\small{0.35}} & \textbf{63.24$\pm$\small{0.52}} & 
    \textbf{41.50$\pm$\small{0.35}} & \textbf{67.81$\pm$\small{0.42}} & 
    \textbf{42.19$\pm$\small{0.28}} & \textbf{68.66$\pm$\small{0.21}} \\
    \bottomrule
  \end{tabular}
 \end{adjustbox}
\end{table}

\subsection{Number of augmentations}\label{appendix:number_augmentations}

\begin{table}[H]
 \caption{Accuracy with respect to the number of augmentations $K$ . Methods have been trained on CIFAR-10 with a Conv-4 backbone for 100 epochs. The input mini-batch has been augmented $K$ times then given as input. Results are the average accuracy (linear evaluation) of three runs on the validation set. Only the relational reasoning accuracy is positively correlated with $K$.}
 \label{tab:number_augmentations}
 \begin{adjustbox}{width=0.9\columnwidth,center}
  \centering
  \begin{tabular}{lcccccc}
    \toprule
    \textbf{Method} &
    $K=2$ & $K=4$ & $K=8$ & $K=16$ & $K=32$\\
    \midrule
    Supervised & 
    79.61$\pm$\small{0.47} & 79.76$\pm$\small{0.54} & 79.96$\pm$\small{0.71} & 79.56$\pm$\small{0.49} &
    80.00$\pm$\small{0.45}\\
    RotationNet \citep{gidaris2018unsupervised} & 
    51.58$\pm$\small{0.49} & 51.51$\pm$\small{1.02} & 52.62$\pm$\small{0.68} & 52.85$\pm$\small{1.24} &
    52.25$\pm$\small{1.06}\\
    \emph{Relational Reasoning} (ours) & 
    55.31$\pm$\small{0.58} & 58.05$\pm$\small{0.67} & 59.24$\pm$\small{0.51} & 60.26$\pm$\small{0.59} &
    60.33$\pm$\small{0.36} \\
    \bottomrule
  \end{tabular}
 \end{adjustbox}
\end{table}

\subsection{Semi-supervised and supervised}\label{appendix:semisupervised}

\begin{table}[H]
 \caption{Test accuracy on \emph{CIFAR-10} with respect to the percentage of labeled data available. Methods have been trained with a ResNet-32 backbone (200 epochs), followed by linear evaluation on the entire labeled dataset (100 epochs). The quality of the representations improves with the number of labeled data available.}
 \label{tab:semisupervised-cifar10}
 \begin{adjustbox}{width=1.0\columnwidth,center}
  \centering
  \begin{tabular}{lccccccc}
    \toprule
    \textbf{Method} &
    $0\%$ & $1\%$ & $10\%$ & $25\%$ & $50\%$& $100\%$\\
    \midrule
    Supervised & 
    n/a & n/a & n/a & n/a & n/a & 90.87$\pm$\small{0.41}\\
    \emph{Relational Reasoning} (ours) & 
    74.99$\pm$\small{0.07} & 76.55$\pm$\small{0.27} & 80.14$\pm$\small{0.35} & 85.30$\pm$\small{0.28} &
    89.35$\pm$\small{0.11} & 90.66$\pm$\small{0.23} \\
    \bottomrule
  \end{tabular}
 \end{adjustbox}
\end{table}

\begin{table}[H]
 \caption{Test accuracy on \emph{CIFAR-100} with respect to the percentage of labeled data available. Methods have been trained with a ResNet-32 backbone (200 epochs), followed by linear evaluation on the entire labeled dataset (100 epochs). The quality of the representations improves with the number of labeled data available.}
 \label{tab:semisupervised-cifar100}
 \begin{adjustbox}{width=1.0\columnwidth,center}
  \centering
  \begin{tabular}{lccccccc}
    \toprule
    \textbf{Method} &
    $0\%$ & $1\%$ & $10\%$ & $25\%$ & $50\%$& $100\%$\\
    \midrule
    Supervised & 
    n/a & n/a & n/a & n/a & n/a & 65.32$\pm$\small{0.22}\\
    \emph{Relational Reasoning} (ours) & 
    46.17$\pm$\small{0.17} & 46.10$\pm$\small{0.29} & 49.55$\pm$\small{0.36} & 54.44$\pm$\small{0.58} &
    58.52$\pm$\small{0.70} & 58.96$\pm$\small{0.28} \\
    \bottomrule
  \end{tabular}
 \end{adjustbox}
\end{table}

\subsection{Ablations} \label{appendix:additional_ablations}
\FloatBarrier

\begin{table}[H]
 \caption{Ablation of the aggregation function. Training with relational self-supervision on unlabeled CIFAR-10 and CIFAR-100, and linear evaluation on labeled datasets (Conv-4). Mean accuracy (percentage) and standard deviation over three runs on a validation set (obtained sampling 20\% of the images from the training set). Best results highlighted in bold.}
 \label{tab:aggregation_comparison}
 \begin{adjustbox}{width=0.65\columnwidth,center}
  \centering
  \begin{tabular}{cccc}
    \toprule
    \textbf{Aggregation} &\textbf{Analytical form} &
    \textbf{CIFAR-10} & \textbf{CIFAR-100} \\
    \midrule
    Sum & $a_{\text{sum}}(\mathbf{z}_i, \mathbf{z}_j) = \mathbf{z}_i + \mathbf{z}_j$ & 
    57.60$\pm$\small{0.23} & 29.45$\pm$\small{0.69} \\
    Mean & $a_{\text{mean}}(\mathbf{z}_i, \mathbf{z}_j) = \frac{\mathbf{z}_i + \mathbf{z}_j}{2}$ & 
    57.77$\pm$\small{0.74} & 29.15$\pm$\small{0.80} \\
    Maximum & $a_{\text{max}}(\mathbf{z}_i, \mathbf{z}_j) = \text{max}(\mathbf{z}_i, \mathbf{z}_j)$ & 
    56.45$\pm$\small{1.15} & 26.58$\pm$\small{1.26} \\
    Concatenation & $a_{\text{cat}}(\mathbf{z}_i, \mathbf{z}_j) =  \mathbf{z}_i \frown \mathbf{z}_j$ 
    & \textbf{60.81$\pm$\small{0.25}} & \textbf{32.36$\pm$\small{0.73}}\\
    \bottomrule
  \end{tabular}
 \end{adjustbox}
\end{table}

\begin{table}[H]
 \caption{Ablation of the relation head. The models have been trained on unlabeled CIFAR-10 and CIFAR-100 and tested on various benchmarks with a ResNet32 backbone for 200 epochs (mean accuracy and standard deviation of 3 runs). We consider three head types: \emph{(a)~dot product} between the pairs encoded through the backbone, followed by BCE loss; \emph{(b)~Encoder + dot product}, aggregation is not performed, for each encoded representation an MLP performs a second encoding, then dot product is applied between pairs and the BCE loss minimized (similar to SimCLR, \citealt{chen2020simple}); \emph{(c)~Relation module} corresponds to the proposed method where encodings are aggregated (concatenation) and passed through an MLP for binary classification. All the other factors are kept constant for a fair comparison (e.g. augmentation strategy, mini-batch size). Best results in bold.}
 \label{tab:head_type}
 \begin{adjustbox}{width=0.75\columnwidth,center}
  \centering
  \begin{tabular}{lcccccc}
    \toprule
     & \multicolumn{2}{c}{\textbf{Linear Evaluation}} & \multicolumn{2}{c}{\textbf{Domain Transfer}} &
     \multicolumn{1}{c}{\textbf{Grain}}\\
    \cmidrule[0.1pt](r){2-3} \cmidrule[0.1pt](r){4-5} \cmidrule[0.1pt](l){6-6}
    \textbf{Head type} &
    \textbf{CIFAR-10} & \textbf{CIFAR-100} & \textbf{10}$\rightarrow$\textbf{100} & \textbf{100}$\rightarrow$\textbf{10} & \textbf{CIFAR-100-20}\\
    \midrule
    (a) dot product & 
    72.74$\pm$\small{0.22} & 28.77$\pm$\small{0.44} & 18.19$\pm$\small{0.10} & 51.9$\pm$\small{0.50} & 45.05$\pm$\small{1.07}\\
    (b) Encoder + dot product & 
    59.44$\pm$\small{0.59} & 29.91$\pm$\small{1.28} & 28.29$\pm$\small{0.90} & 53.65$\pm$\small{0.85} & 36.94$\pm$\small{1.30}\\
    (c) \emph{Relation module} (ours) & 
    \textbf{74.99$\pm$\small{0.07}} & \textbf{46.17$\pm$\small{0.17}} & \textbf{41.50$\pm$\small{0.35}} & \textbf{67.81$\pm$\small{0.42}} & \textbf{52.44$\pm$\small{0.47}}\\
    \bottomrule
  \end{tabular}
 \end{adjustbox}
\end{table}

\begin{figure}[H]
    \centering
    \begin{subfigure}[t]{0.30\textwidth}
       \centering
        \includegraphics[width=0.75\textwidth, trim={0.0cm 0.0cm 0.0cm 0.0cm}, clip]{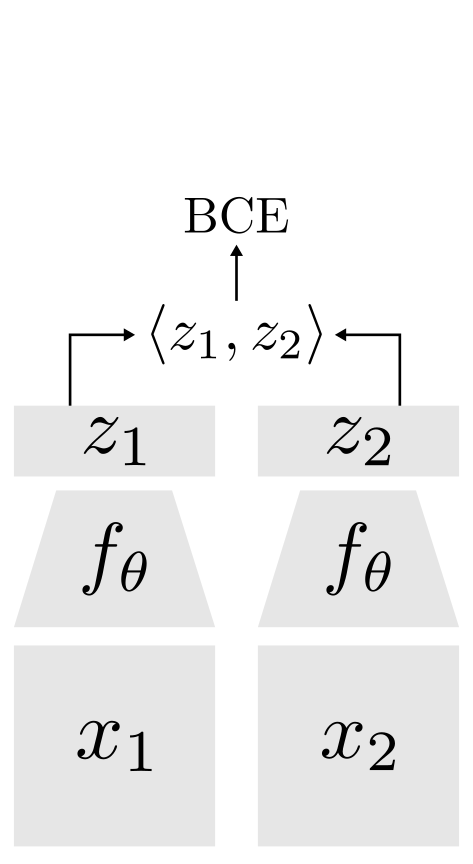}
        \caption{dot product}
        \label{fig:ablation-nohead}
    \end{subfigure}
    \begin{subfigure}[t]{0.30\textwidth}
       \centering
        \includegraphics[width=0.75\textwidth, trim={0.0cm 0.0cm 0.0cm 0.0cm}, clip]{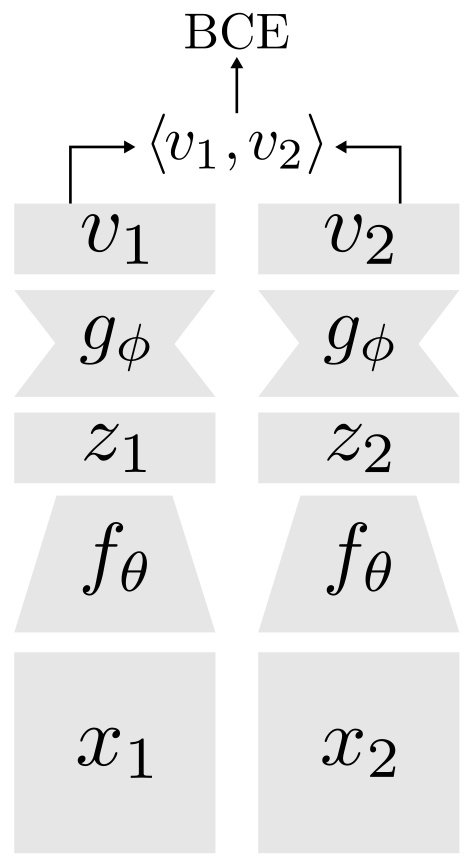}
        \caption{Encoder + dot product}
        \label{fig:ablation-encoder}
    \end{subfigure}
    \begin{subfigure}[t]{0.30\textwidth}
       \centering
        \includegraphics[width=0.75\textwidth, trim={0.0cm 0.0cm 0.0cm 0.0cm}, clip]{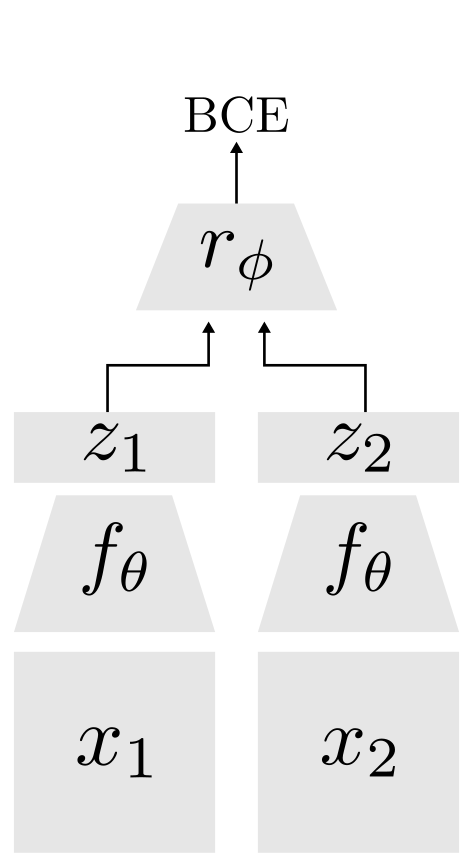}
        \caption{Relation module (ours)}
        \label{fig:ablation-ours}
    \end{subfigure}
    \caption{Ablation of the relation head (graphical illustration). Comparison between the two ablations in (\subref{fig:ablation-nohead}) and (\subref{fig:ablation-encoder}), and the full model in (\subref{fig:ablation-ours}). In (\subref{fig:ablation-nohead}) the head is removed and the dot product $\langle z_{1}, z_{2} \rangle$ is used to compare the representations pair. In (\subref{fig:ablation-encoder}) the relation head is replaced with an encoder $g_{\phi}$ that projects each representation in another latent space where the dot product is performed. In (\subref{fig:ablation-ours}) is showed the full model, with the relation module $r_{\phi}$ taking in input the aggregated pair. In all cases is minimized the binary cross-entropy loss (BCE) over positive and negative pairs.}
    \label{fig:ablation}
\end{figure}

\subsection{Image retrieval: qualitative analysis}\label{appendix:additional_retrieval}
\FloatBarrier

\begin{figure}[H]
    \begin{subfigure}[t]{0.5\textwidth}
        \centering
        \includegraphics[width=0.95\textwidth, trim={0.0cm 0.0cm 0.0cm 0.0cm}, clip]{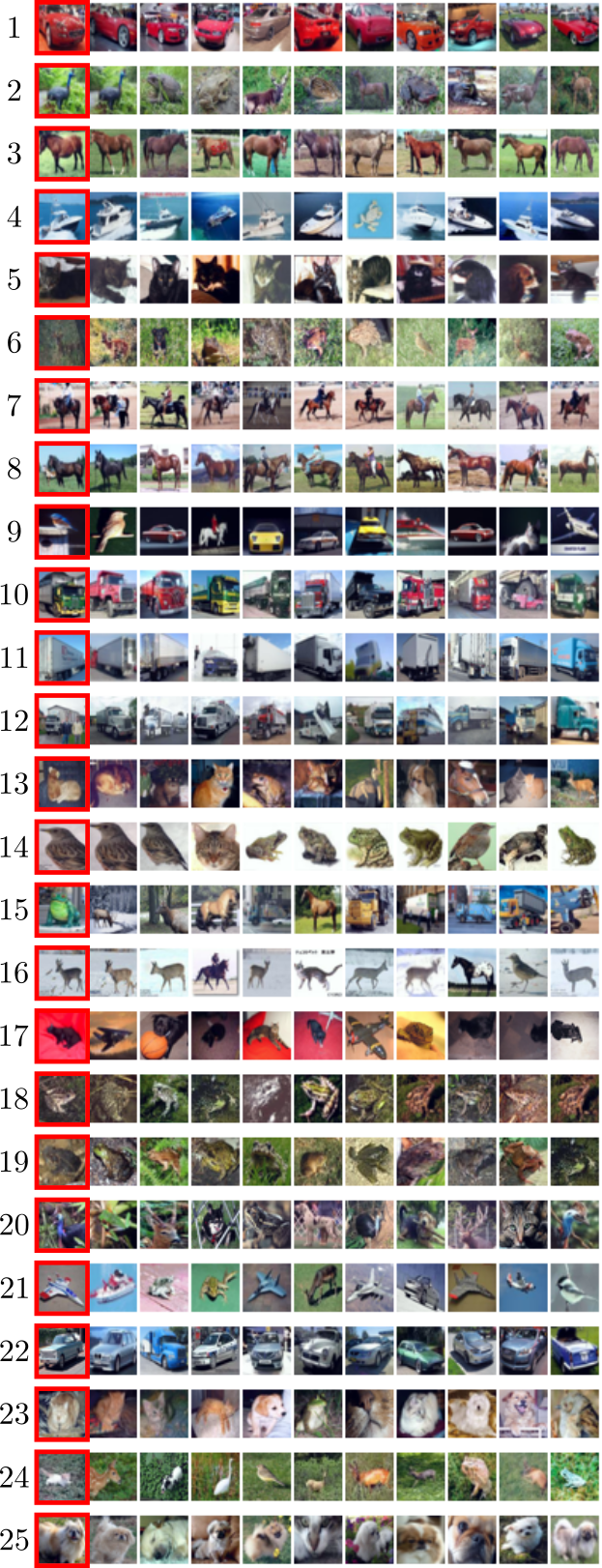}
        \caption{Relational Reasoning (ours)}
        \label{fig:retrieval-relation}
    \end{subfigure}
    \begin{subfigure}[t]{0.5\textwidth}
        \centering
        \includegraphics[width=0.95\textwidth, trim={0.0cm 0.0cm 0.0cm 0.0cm}, clip]{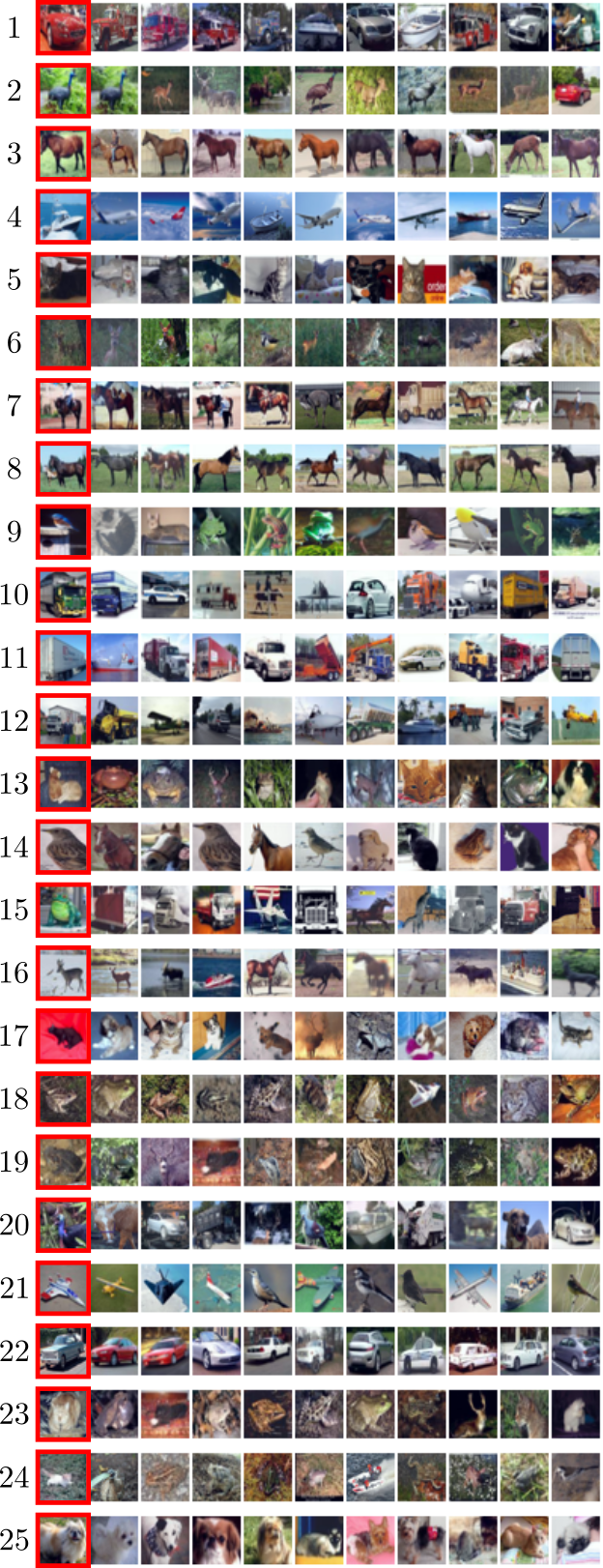}
        \caption{RotationNet}
        \label{fig:retrieval-rotation}
    \end{subfigure}
    \caption{Image retrieval given 25 random queries (not cherry-picked) on CIFAR-10 with ResNet-32. The query is the leftmost image (red frame), followed by the top-10 most similar images (Euclidean distance) in representation space. Comparison between (\subref{fig:retrieval-relation}) self-supervised relational reasoning (ours), and (\subref{fig:retrieval-rotation}) self-supervised rotation prediction \citep{gidaris2018unsupervised}. Our method shows better distinction between categories which are hard to separate, (e.g. ships vs planes in row 4, trucks vs cars in row 12). Moreover, the lower sample variance and the higher similarity with the query, indicates a fine-grained organization in representation space (e.g. red sport cars in row 1, long white trucks in rows 11 and 12, deer with snow in row 16, blue car in row 22, dog breeds in row 25).}
    \label{fig:retrieval}
\end{figure}

\subsection{Image retrieval: error analysis}\label{appendix:additional_retrieval_error}
\FloatBarrier

\begin{figure}[H]
    \begin{subfigure}[t]{0.5\textwidth}
        \centering
        %left, bottom, right, top
        \includegraphics[width=0.95\textwidth, trim={1.3cm 0.3cm 3.1cm 0.9cm}, clip]{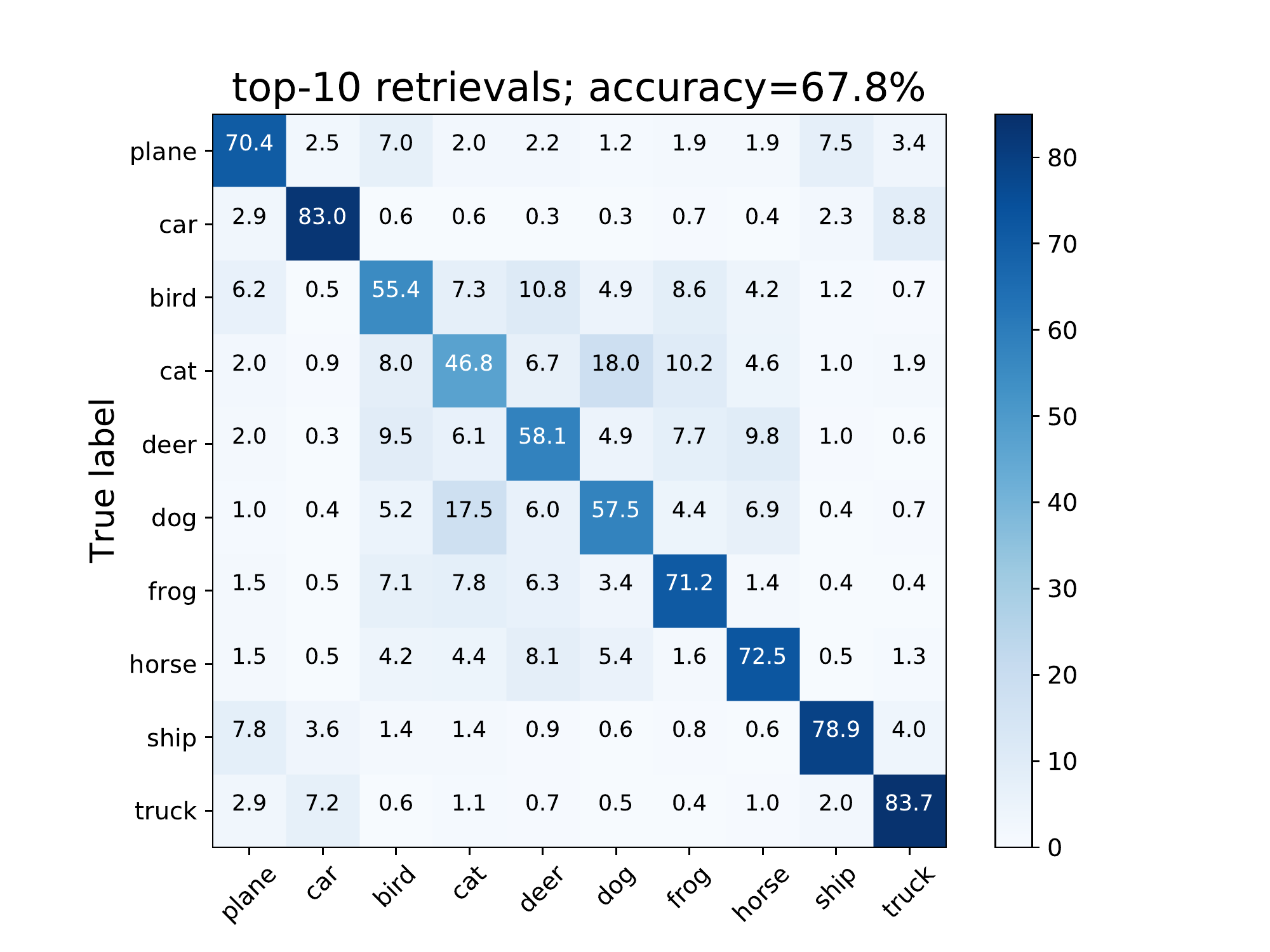}
        \includegraphics[width=0.95\textwidth, trim={1.3cm 0.3cm 3.1cm 0.9cm}, clip]{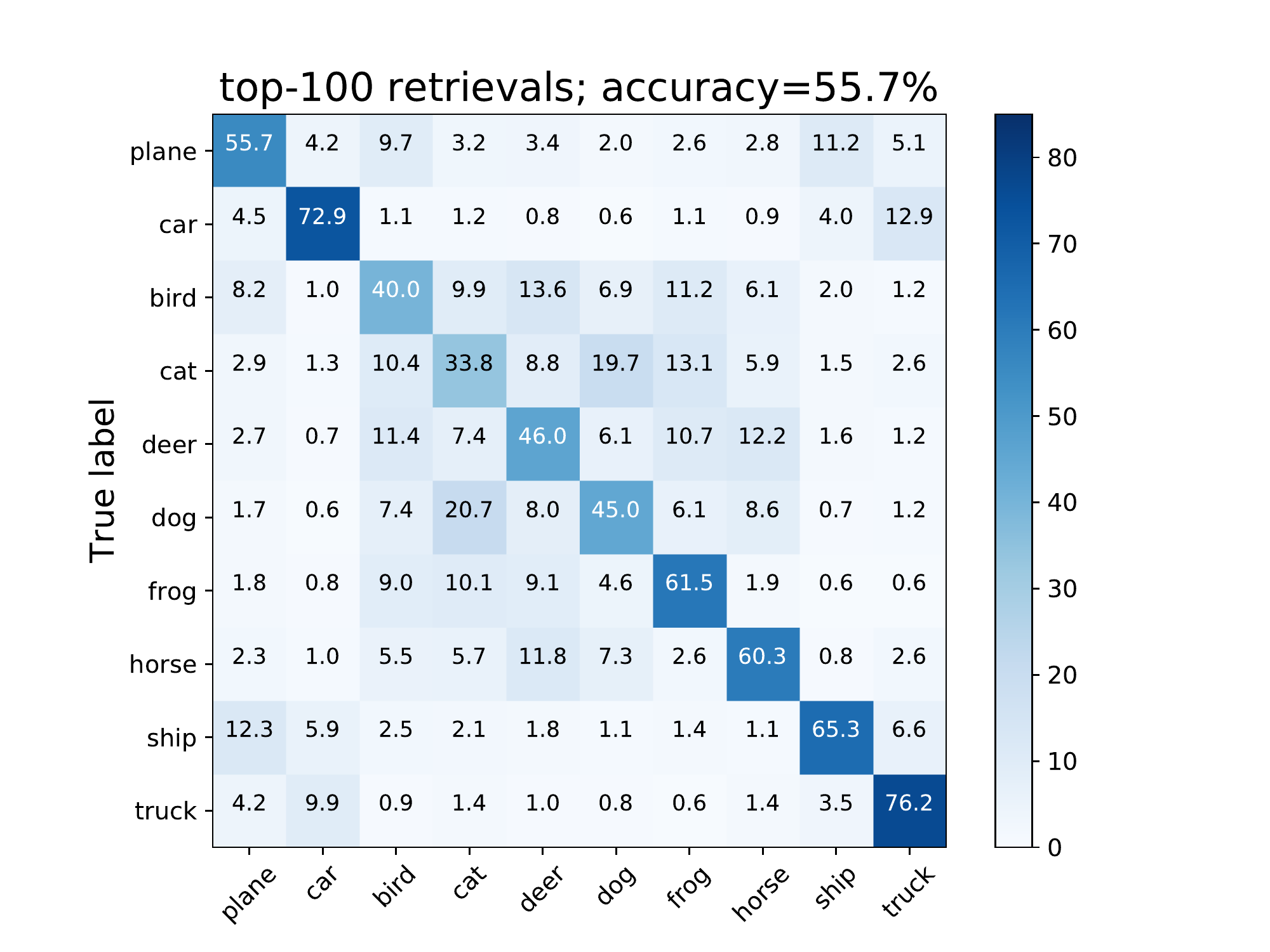}
        \includegraphics[width=0.95\textwidth, trim={1.3cm 0.3cm 3.1cm 0.9cm}, clip]{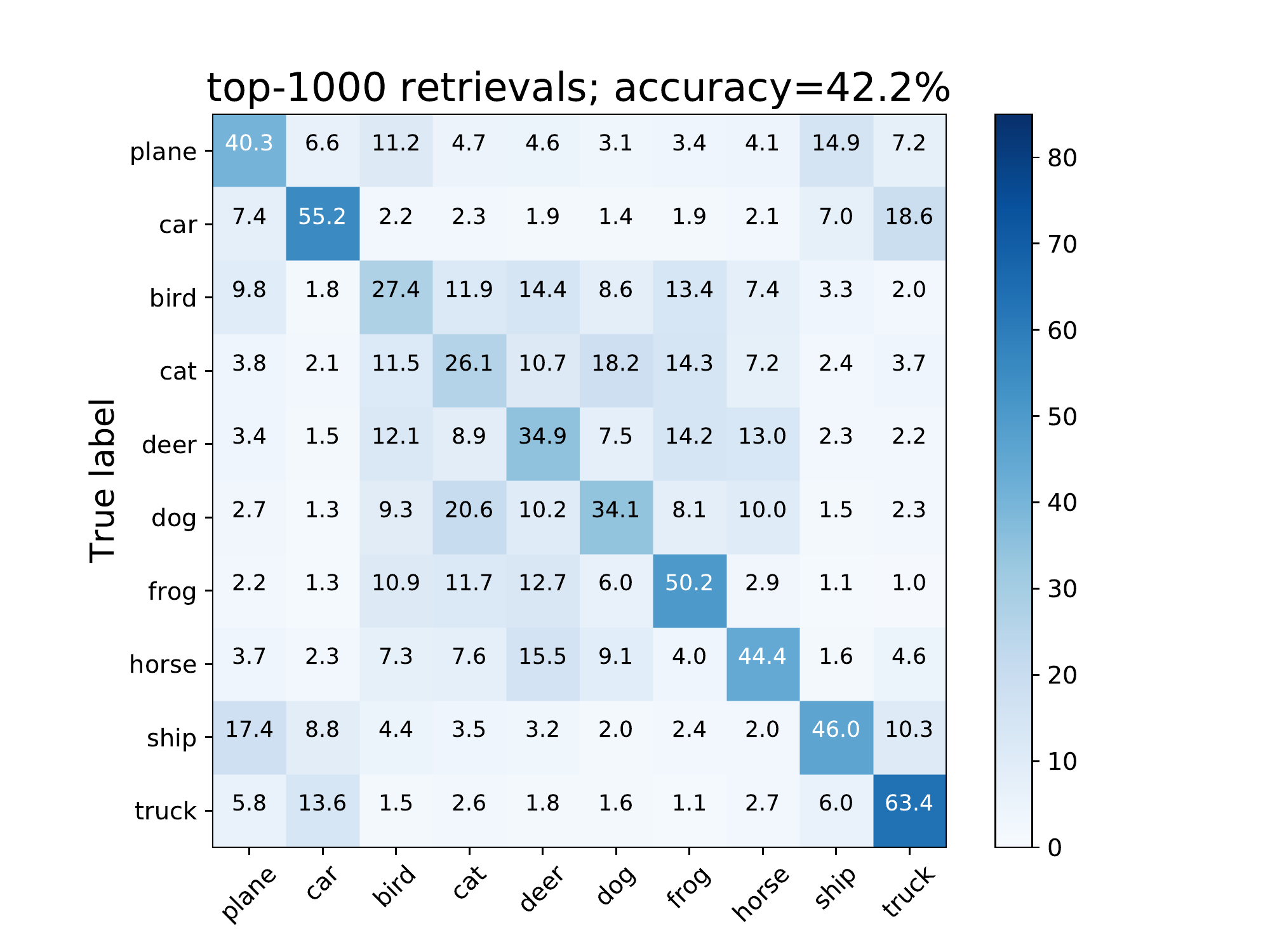} 
        \caption{Relational Reasoning (ours)}
        \label{fig:retrieval-error-relation}
    \end{subfigure}
    \begin{subfigure}[t]{0.5\textwidth}
        \centering
        \includegraphics[width=0.95\textwidth, trim={1.3cm 0.3cm 3.1cm 0.9cm}, clip]{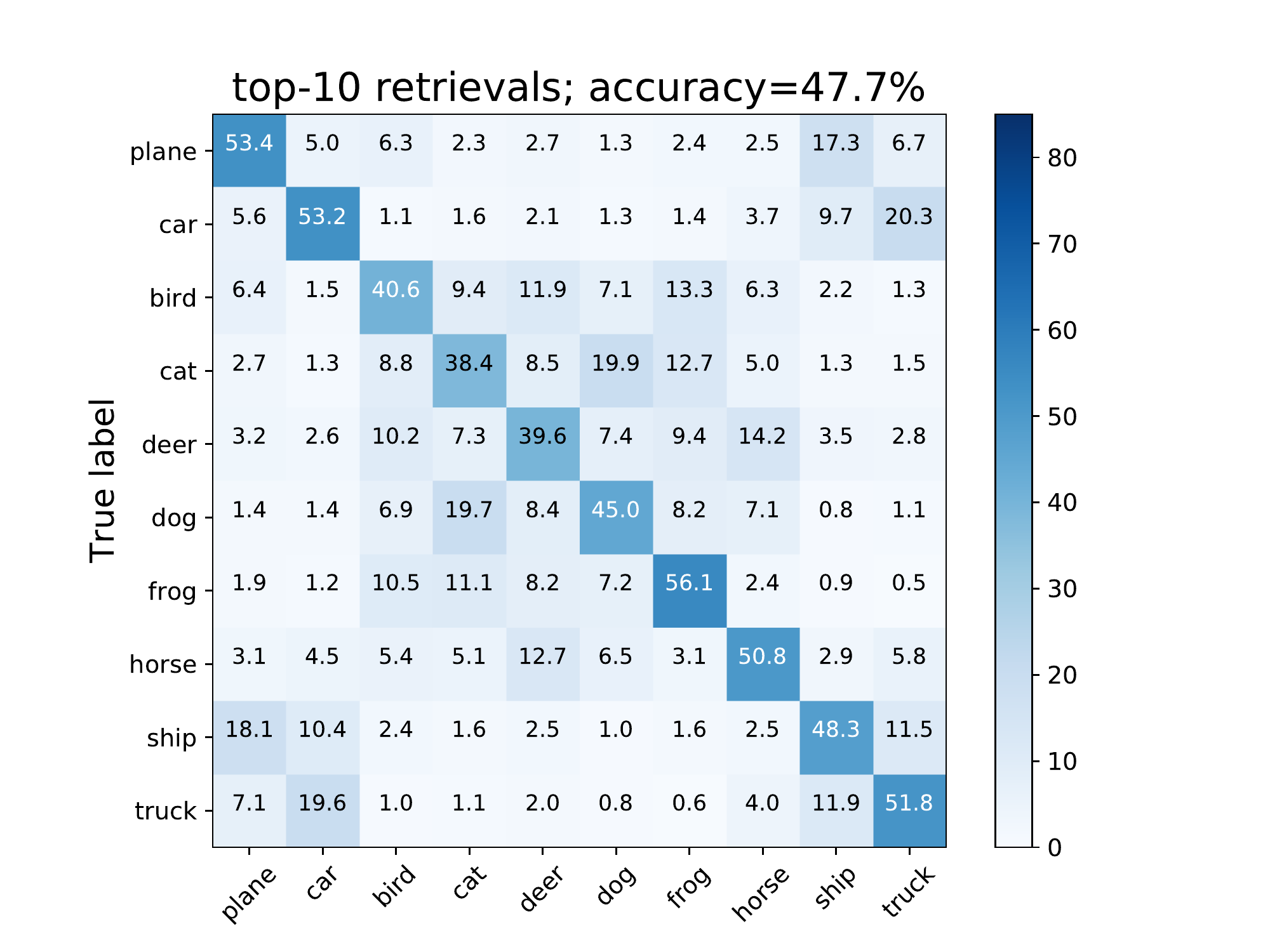}
        \includegraphics[width=0.95\textwidth, trim={1.3cm 0.3cm 3.1cm 0.9cm}, clip]{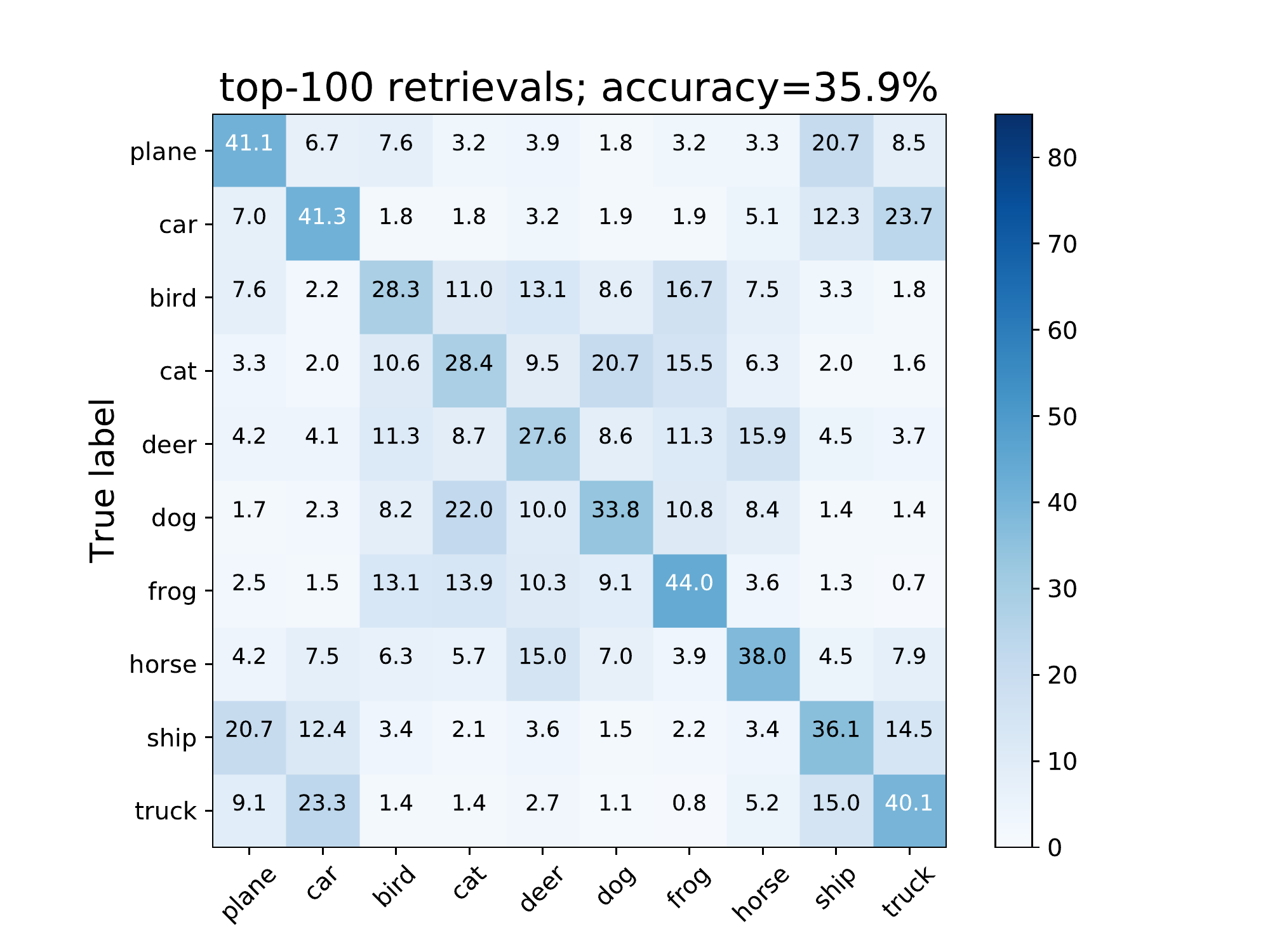}
        \includegraphics[width=0.95\textwidth, trim={1.3cm 0.3cm 3.1cm 0.9cm}, clip]{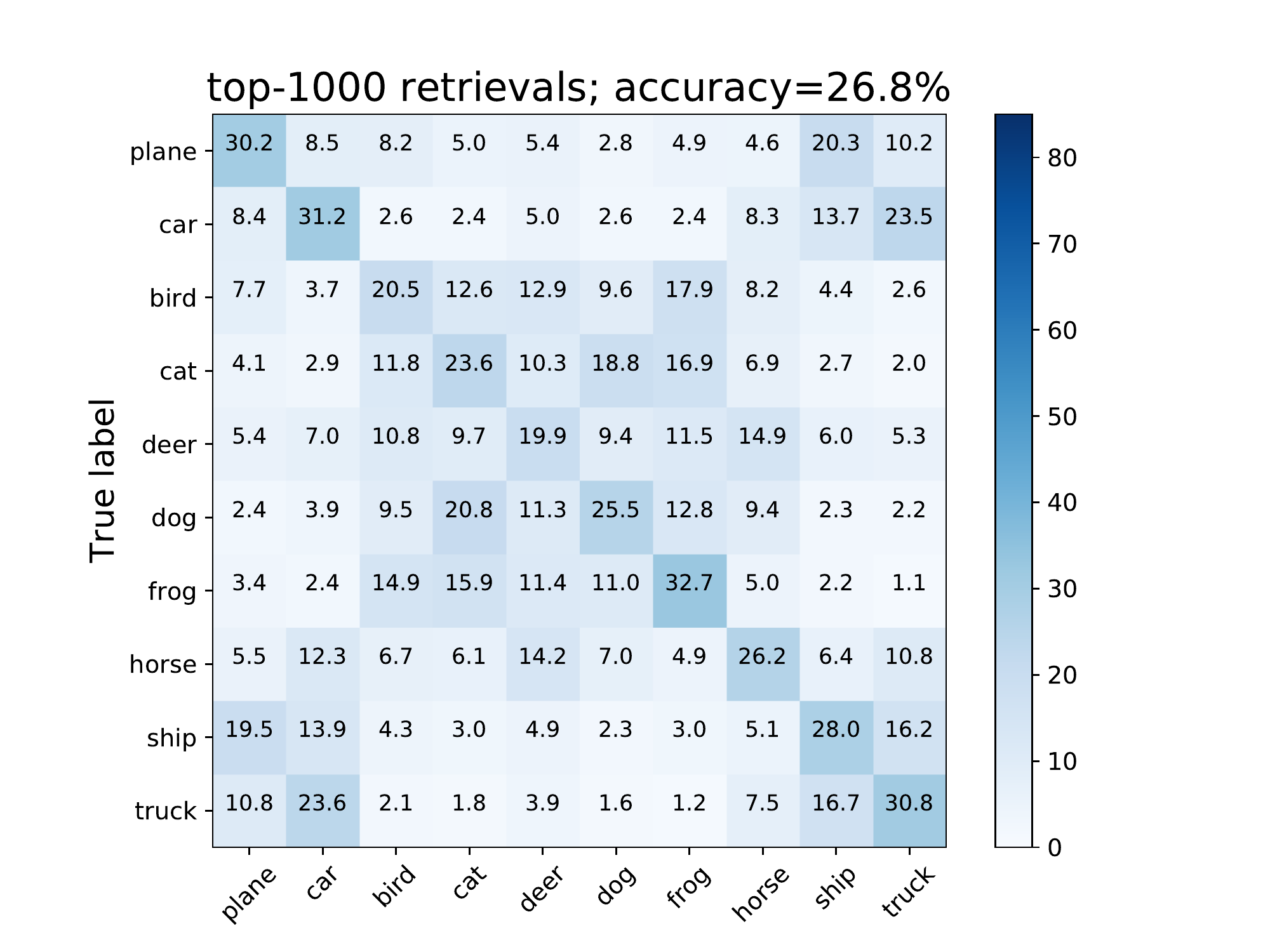}
        \caption{RotationNet}
        \label{fig:retrieval-error-rotation}
    \end{subfigure}
    \caption{Confusion matrix obtained sampling 500 images per class (CIFAR-10, ResNet-32) and retrieving the top-10/100/1000 (top/middle/bottom table) closest images in representation space via Euclidean distance. Accuracy in percentage (three seeds) over correct retrievals (same category). Comparison between (\subref{fig:retrieval-error-relation}) self-supervised relational reasoning (ours), and (\subref{fig:retrieval-error-rotation}) self-supervised rotation prediction \citep{gidaris2018unsupervised}. The proposed method shows a superior accuracy across all categories while being more robust against misclassification errors.}
    \label{fig:retrieval-error}
\end{figure}

\subsection{Representations: qualitative analysis}\label{appendix:additional_qualitative}
\FloatBarrier

\begin{figure}[H]
    \begin{subfigure}[t]{0.333\textwidth}
       \centering
        \includegraphics[width=1.0\textwidth, trim={0.5cm 0.45cm 0.25cm 0.45cm}, clip]{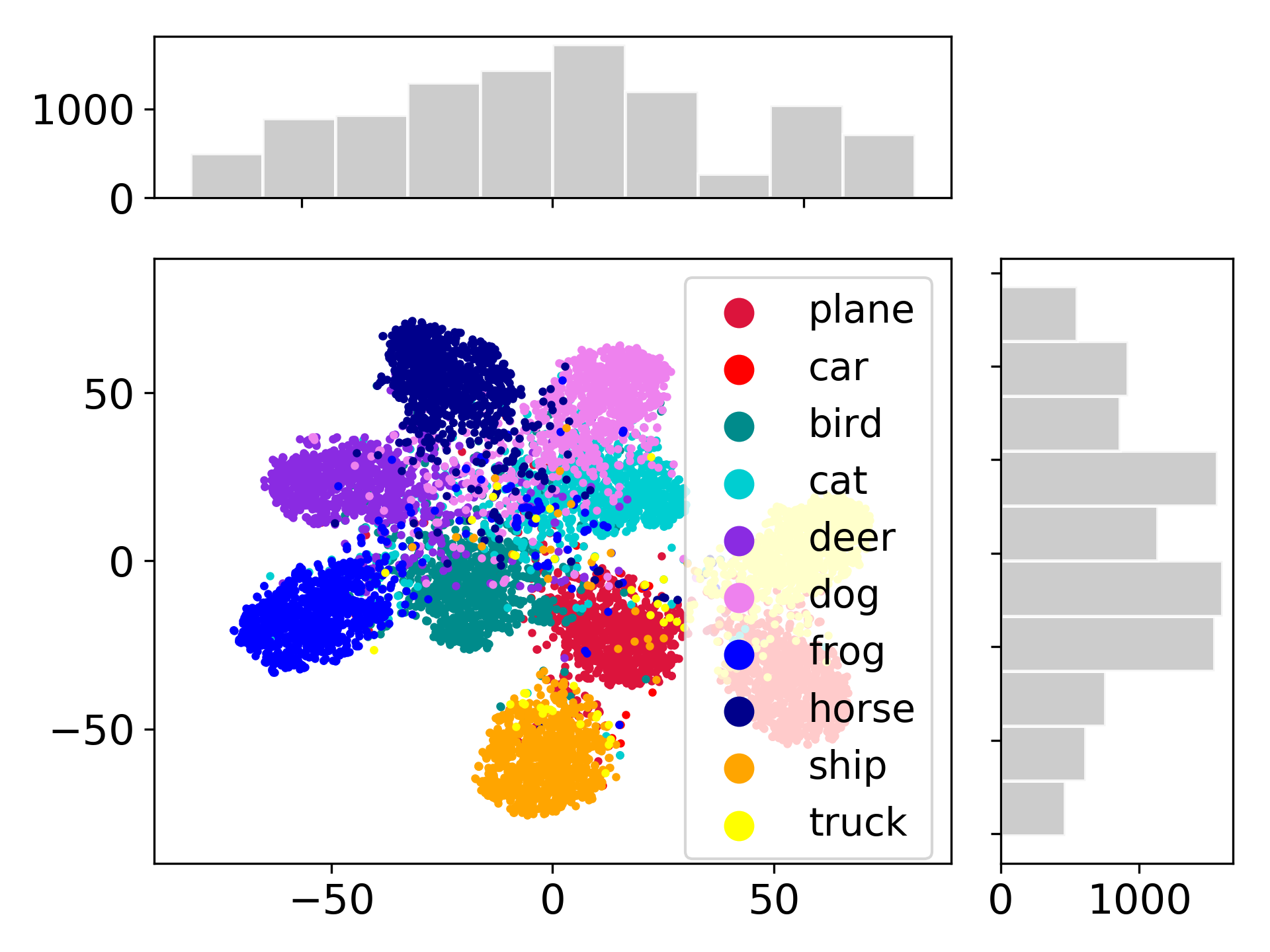}
        \caption{Supervised}
        \label{fig:tsne-supervised}
    \end{subfigure}
    \begin{subfigure}[t]{0.333\textwidth}
       \centering
        \includegraphics[width=1.0\textwidth, trim={0.5cm 0.45cm 0.25cm 0.45cm}, clip]{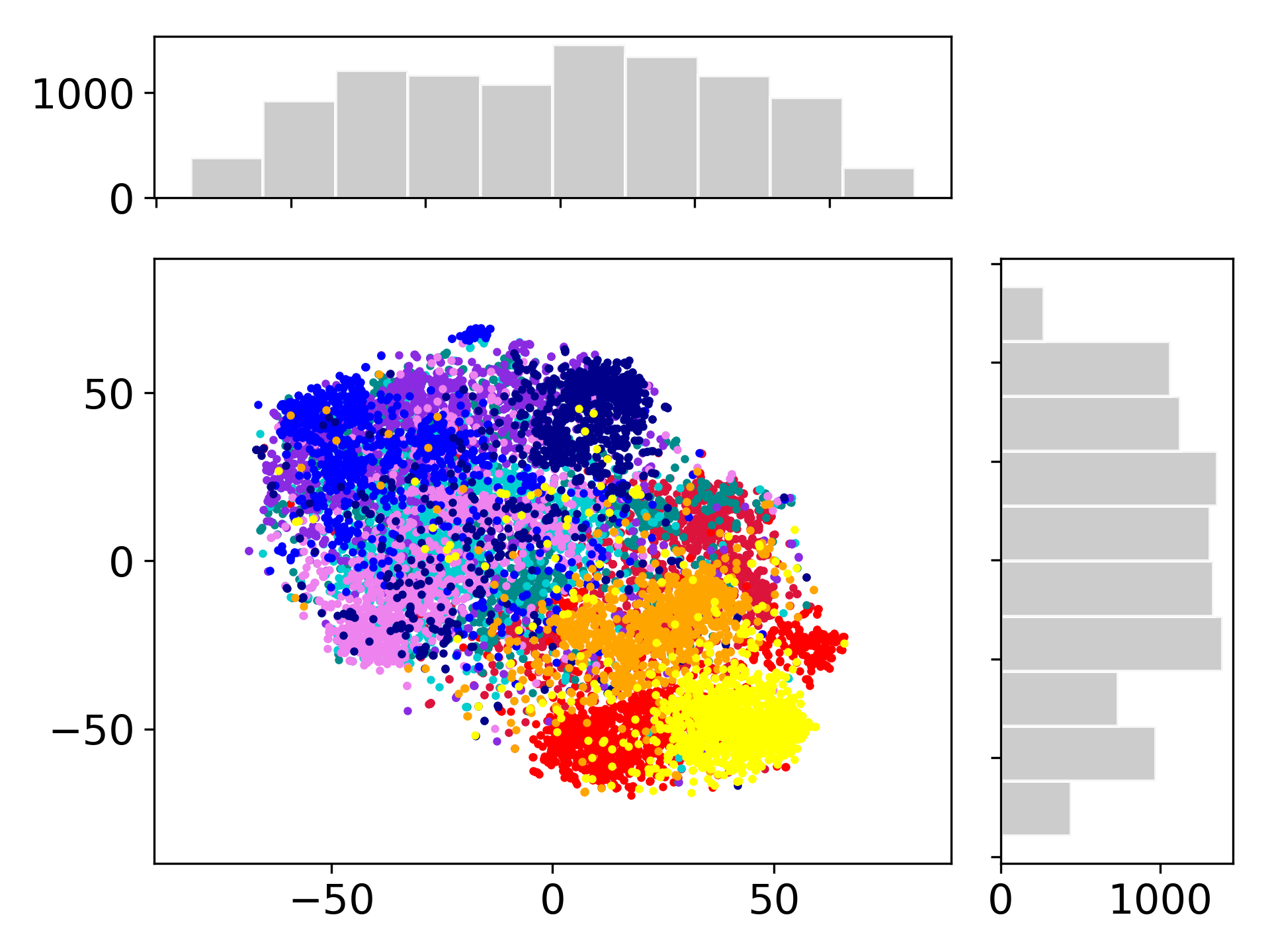}
        \caption{Relational Reasoning (ours)}
        \label{fig:tsne-relationat}
    \end{subfigure}
    \begin{subfigure}[t]{0.333\textwidth}
       \centering
        \includegraphics[width=1.0\textwidth, trim={0.5cm 0.45cm 0.25cm 0.45cm}, clip]{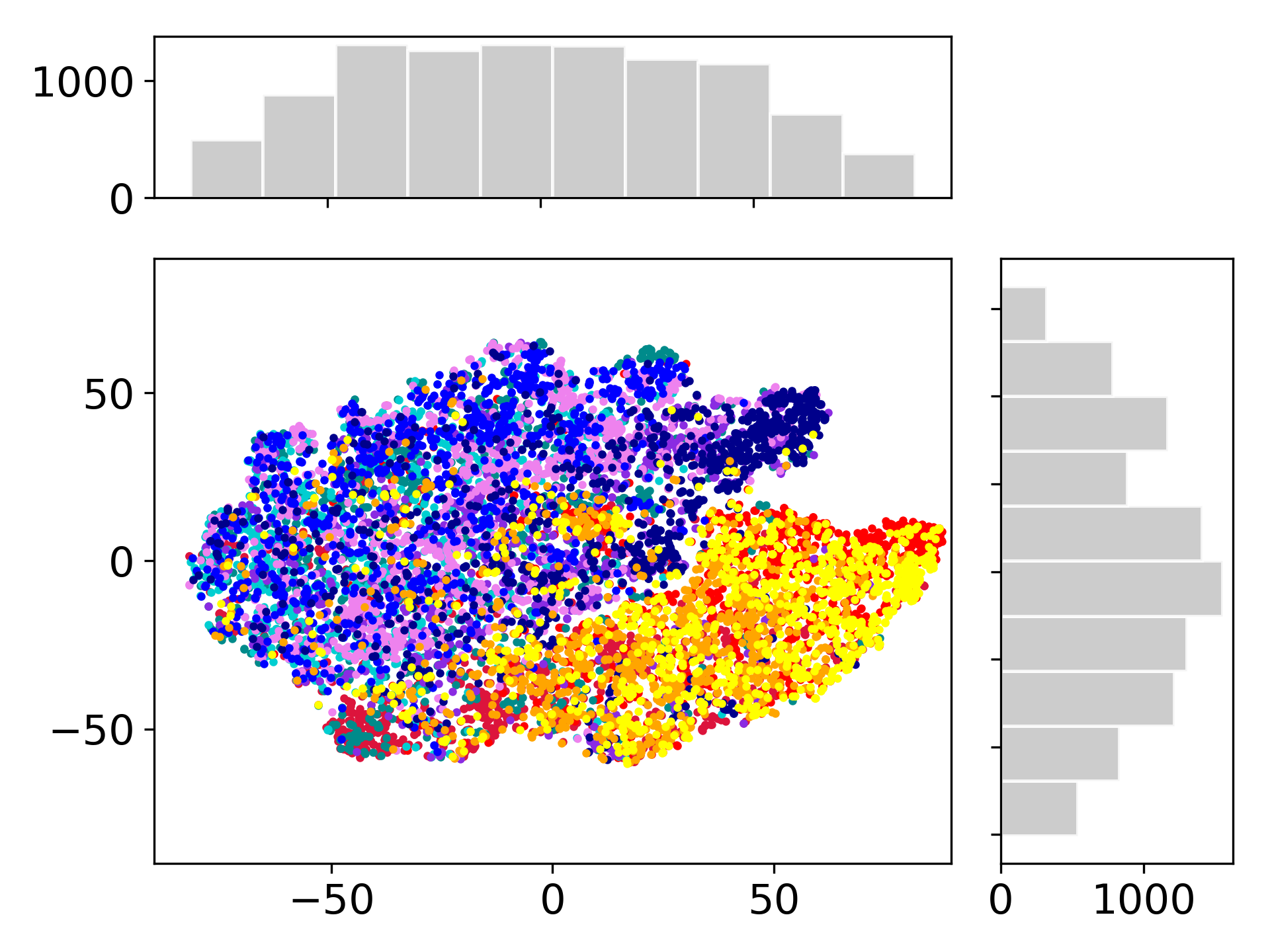}
        \caption{RotationNet}
        \label{fig:tsne-rotation}
    \end{subfigure}
    \caption{Visualization of t-SNE embeddings for the 10K test points in CIFAR-10. ResNet-32 backbone trained via (\subref{fig:tsne-supervised}) supervised learning, (\subref{fig:tsne-relationat}) self-supervised relational reasoning (ours), and (\subref{fig:tsne-rotation}) self-supervised rotation prediction \citep{gidaris2018unsupervised}. Our method shows a lower scattering, with clusters which are more distinct.}
    \label{fig:tsne}
\end{figure}

\begin{figure}[H]
    \begin{subfigure}[t]{0.333\textwidth}
        \centering
        \includegraphics[width=1.0\textwidth, trim={0.5cm 0.45cm 0.25cm 0.45cm}, clip]{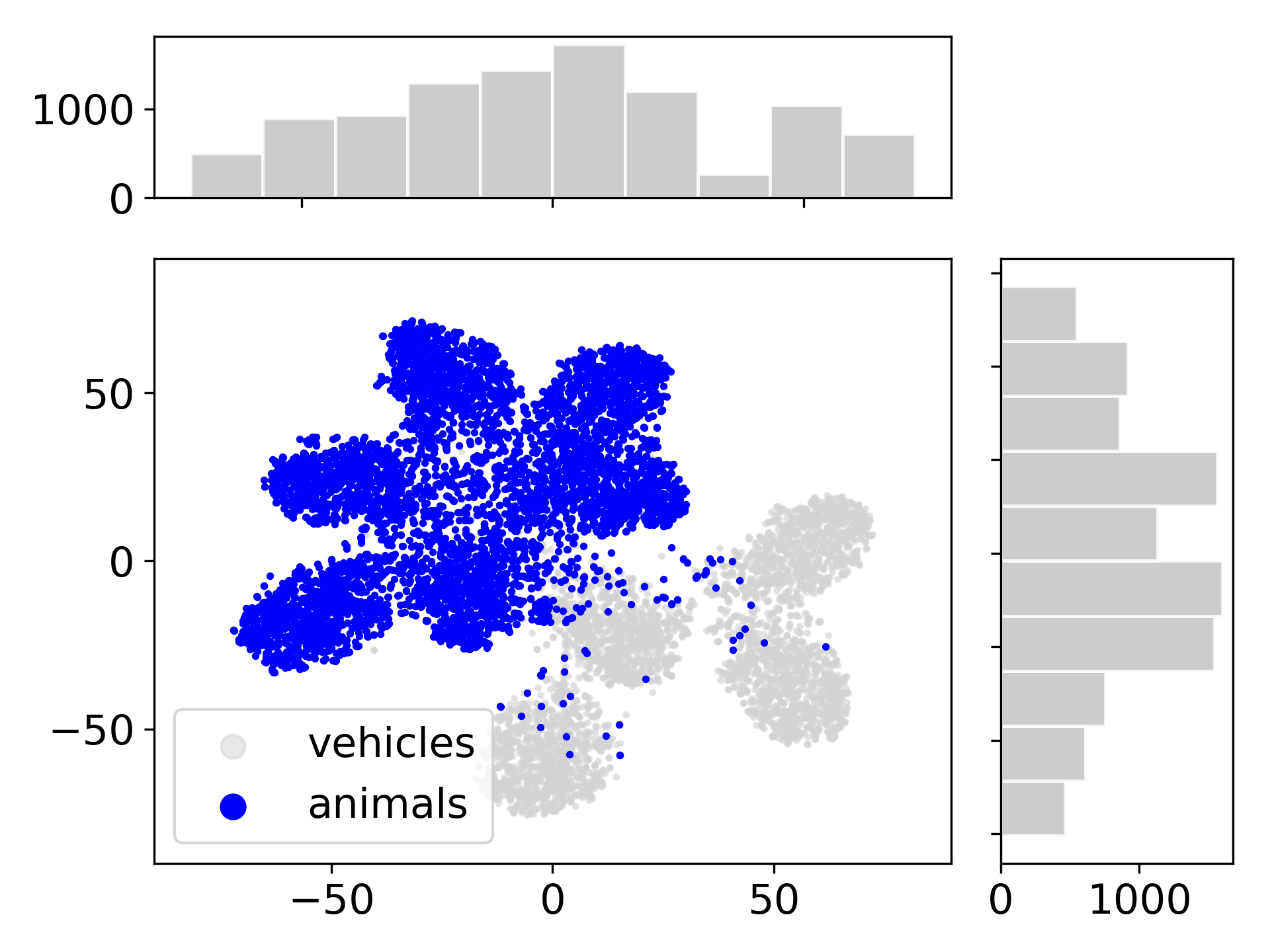}
    \end{subfigure}
    \begin{subfigure}[t]{0.333\textwidth}
       \centering
        \includegraphics[width=1.0\textwidth, trim={0.5cm 0.45cm 0.25cm 0.45cm}, clip]{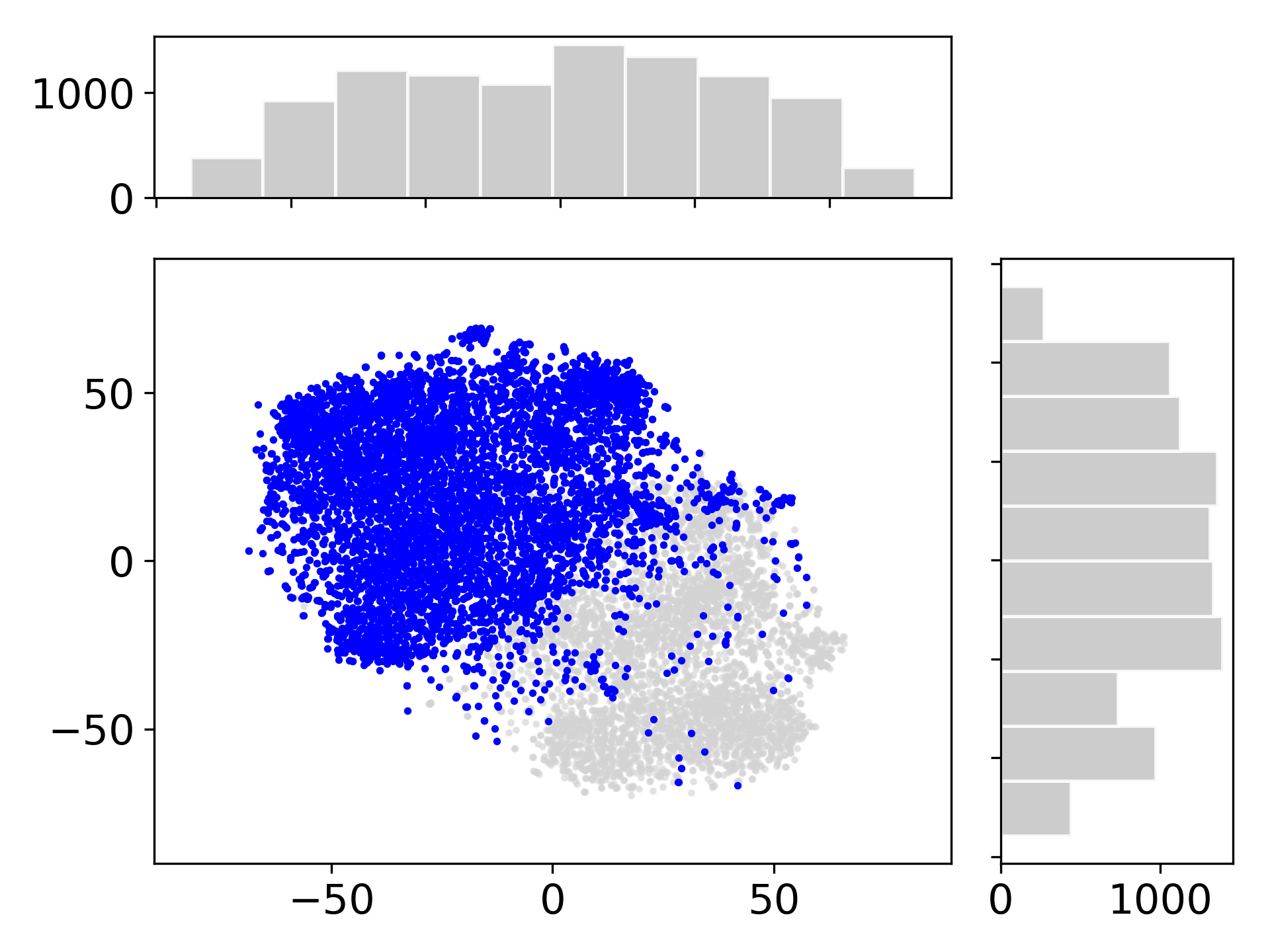}
    \end{subfigure}
    \begin{subfigure}[t]{0.333\textwidth}
       \centering
        \includegraphics[width=1.0\textwidth, trim={0.5cm 0.45cm 0.25cm 0.45cm}, clip]{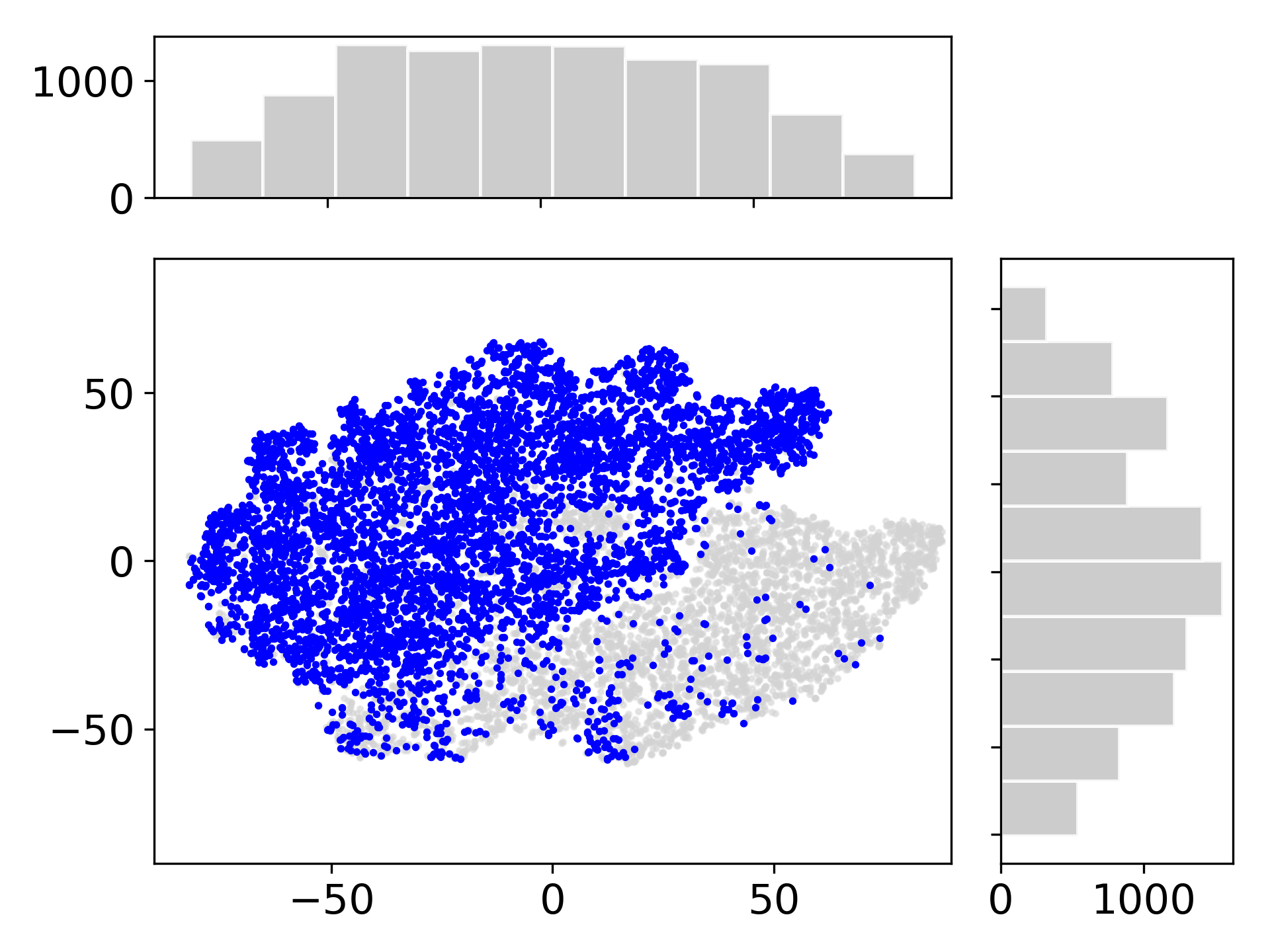}
    \end{subfigure}
    \begin{subfigure}[t]{0.333\textwidth}
        \centering
        \includegraphics[width=1.0\textwidth, trim={0.5cm 0.45cm 0.25cm 2.8cm}, clip]{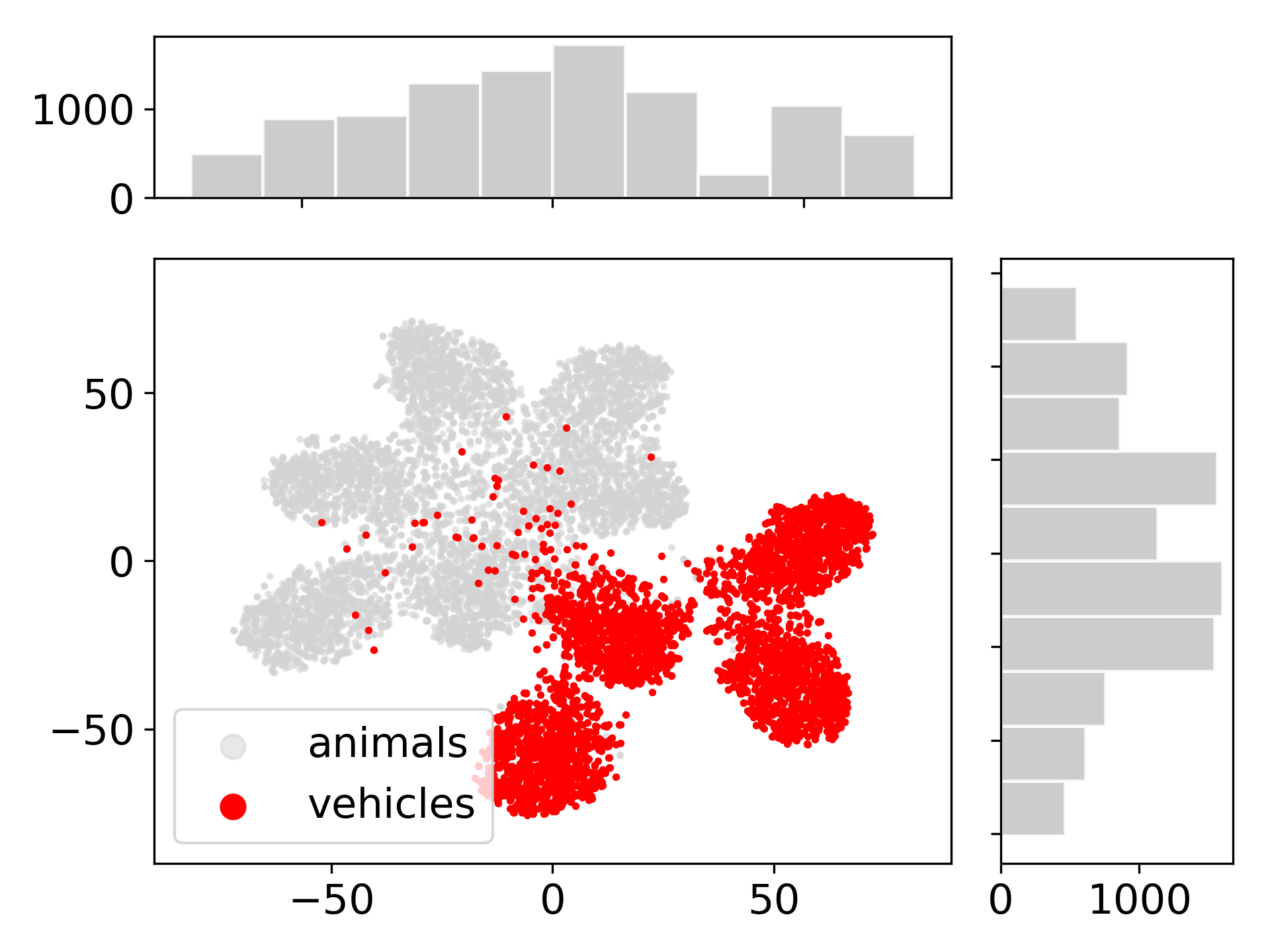}
        \caption{Supervised}
        \label{fig:tsne-va-supervised}
    \end{subfigure}
    \begin{subfigure}[t]{0.333\textwidth}
        \centering
        \includegraphics[width=1.0\textwidth, trim={0.5cm 0.45cm 0.25cm 2.8cm}, clip]{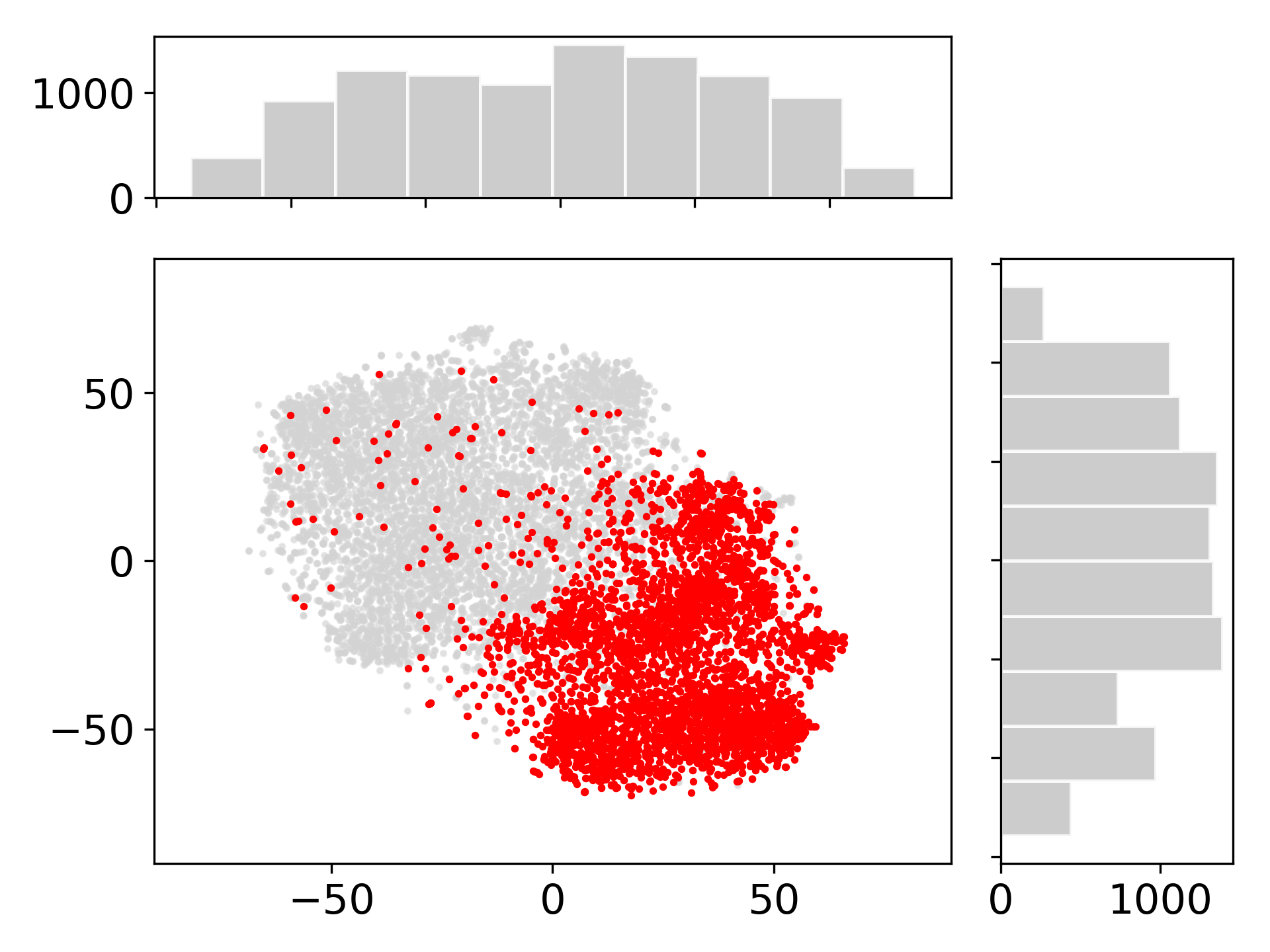}
        \caption{Relational Reasoning (ours)}
        \label{fig:tsne-va-relationat}
    \end{subfigure}%
    \begin{subfigure}[t]{0.333\textwidth}
        \centering
        \includegraphics[width=1.0\textwidth, trim={0.5cm 0.45cm 0.25cm 2.8cm}, clip]{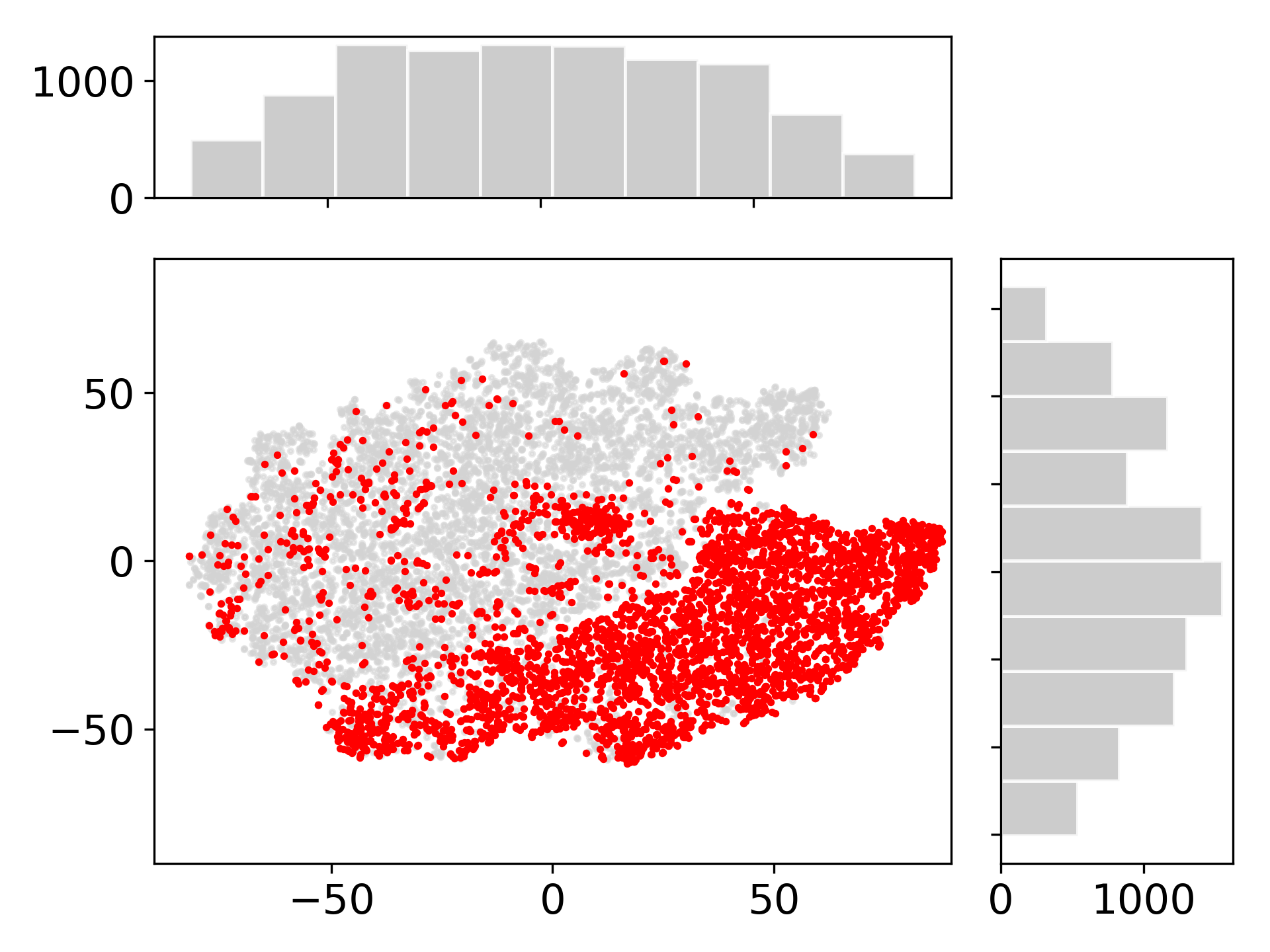}
        \caption{RotationNet}
        \label{fig:tsne-va-rotation}
    \end{subfigure}    
    \caption{Visualization of t-SNE embeddings for the 10K test points in CIFAR-10 divided in two super-categories: vehicles (plane, car, ship, truck), and animals (bird, cat, deer, dog, frog, horse). ResNet-32 backbone trained via (\subref{fig:tsne-va-supervised}) supervised learning, (\subref{fig:tsne-va-relationat}) self-supervised relational reasoning (ours), and (\subref{fig:tsne-va-rotation}) self-supervised rotation prediction \citep{gidaris2018unsupervised}. Our method shows a better split, lower scattering, and a minor overlap between the two super-categories.}
    \label{fig:tsne-va}
\end{figure}

\section{Pseudo-code of the method}\label{appendix:code}
\FloatBarrier

\begin{algorithm}[H]
\caption{Self-supervised relational learning: training function and shuffling without collisions.}
\label{alg:overview}
\textbf{Require:}  $\mathcal{D} = \{\mathbf{x}_n\}_{n=1}^{N}$ unlabeled training set; $\mathcal{A}(\cdot)$ augmentation distribution; $\boldsymbol{\theta}$ parameters of $f_{\theta}$ (neural network backbone); $\boldsymbol{\phi}$ parameters of $r_{\phi}$ (relation module); aggregation function $a(\cdot, \cdot)$; $\alpha$ and $\beta$ learning rate hyperparameters; $K$ number of augmentations; $M$ mini-batch size;\\
\begin{algorithmic}[1]
\Function{Train}{$\mathcal{D}$, $\alpha$, $\beta$, $M$, $K$, $\boldsymbol{\theta}$, $\boldsymbol{\phi}$}
\While{not done}
    \State $\mathcal{B}=\{\mathbf{x}_m\}_{m=1}^{M} \sim \mathcal{D}$ \Comment Sampling a mini-batch
    \For{$k = 1$ to $K$}
        \State $\mathcal{B}^{(k)} \sim \mathcal{A}(\mathcal{B}) $ \Comment Sampling $K$ mini-batch augmentations
        \State $\mathcal{Z}^{(k)} = f_{\theta}(\mathcal{B}^{(k)})$ \Comment Forward pass in the backbone
    \EndFor
    \State $\mathcal{P} = \{ \}$ \Comment Empty set to store aggregated pairs and targets
    \For{$i = 1$ to $K-1$}
        \For{$j = i+1$ to $K$}
            \State $\mathcal{P} \leftarrow \big( a(\mathcal{Z}^{(i)}, \mathcal{Z}^{(j)}), \mathbf{t}=\boldsymbol{1} \big)$ \Comment Aggregating and appending positive pairs
            \State $\tilde{\mathcal{Z}}^{(j)} =$ \Call{Shuffle}{$\mathcal{Z}^{(j)}$} \Comment Shuffling without collisions
            \State $\mathcal{P} \leftarrow \big( a(\mathcal{Z}^{(i)}, \tilde{\mathcal{Z}}^{(j)}), \mathbf{t}=\boldsymbol{0} \big)$ \Comment Aggregating and appending negative pairs
       \EndFor 
    \EndFor
    \State $\mathbf{y} = r_{\phi}(\mathcal{P})$ \Comment Forward pass in the relation module
    \State $\mathcal{L}=\text{BCE}(\mathbf{y}, \mathbf{t})$ \Comment Estimating the Binary Cross-Entropy loss
    \State $\boldsymbol{\theta} \leftarrow \boldsymbol{\theta} - \alpha \nabla_{\theta} \mathcal{L}$ \Comment Updating backbone
    \State $\boldsymbol{\phi} \leftarrow \boldsymbol{\phi} - \beta \nabla_{\phi} \mathcal{L}$ \Comment Updating relation module
\EndWhile
\State \Return{$\boldsymbol{\theta}, \boldsymbol{\phi}$} \Comment Returning the learned weights
\EndFunction
\vspace{0.25cm} %\item[] %empty line
\Function{Shuffle}{$\mathcal{Z}$}
    \State $\tilde{\mathcal{Z}} = \mathcal{Z}$ \Comment Copying the input set
    \For{$m = 1$ to $M$}
        \State $\tilde{m} \sim \{1, \ldots, M \} \setminus \{m\}$ \Comment Sampling an index $\tilde{m} \neq m$
        \State $\tilde{\mathcal{Z}}_{m} \leftarrow \mathcal{Z}_{\tilde{m}}$ \Comment Assigning a random representation with index $\tilde{m}$
    \EndFor
    \State \Return{$\tilde{\mathcal{Z}}$} \Comment Returning the shuffled set
\EndFunction
\Statex
\end{algorithmic}
  \vspace{-0.2cm}%
\end{algorithm}

\section{Essential PyTorch code of the method}\label{appendix:code_python}
\FloatBarrier

%import file with code style
\input{./code/code_style.tex}

\subsection{Data loader}

\begin{lstlisting}[language=Python,numbers=none]
import torchvision
from PIL import Image

class MultiCIFAR10(torchvision.datasets.CIFAR10):
  """Override torchvision CIFAR10 for multi-image management.
     Similar class can be defined for other datasets (e.g. CIFAR100).
  """
  def __init__(self, K, **kwds):
    super().__init__(**kwds)
    self.K = K # tot number of augmentations
            
  def __getitem__(self, index):
    img, target = self.data[index], self.targets[index]
    pic = Image.fromarray(img)            
    img_list = list()
    if self.transform is not None:
      for _ in range(self.K):
        img_transformed = self.transform(pic.copy())
        img_list.append(img_transformed)
    else:
        img_list = img
    return img_list, target
\end{lstlisting}

\subsection{Augmentations}

\begin{lstlisting}[language=Python,numbers=none]
import torchvision.transforms as transforms

normalize = transforms.Normalize(mean=[0.491, 0.482, 0.447], 
                                 std=[0.247, 0.243, 0.262]) # CIFAR10
color_jitter = transforms.ColorJitter(brightness=0.8, contrast=0.8, 
                                      saturation=0.8, hue=0.2)
rnd_color_jitter = transforms.RandomApply([color_jitter], p=0.8)
rnd_gray = transforms.RandomGrayscale(p=0.2)
rnd_rcrop = transforms.RandomResizedCrop(size=32, scale=(0.08, 1.0), 
                                         interpolation=2)
rnd_hflip = transforms.RandomHorizontalFlip(p=0.5)
train_transform = transforms.Compose([rnd_rcrop, rnd_hflip,
                                      rnd_color_jitter, rnd_gray, 
                                      transforms.ToTensor(), normalize
                                      ])
\end{lstlisting}

\subsection{Self-supervised relational reasoning}

\begin{lstlisting}[language=Python,numbers=none]
import torch

class RelationalReasoning(torch.nn.Module):
  def __init__(self, backbone):
    super(RelationalReasoning, self).__init__()
    feature_size = 64*2 # multiply by 2 since aggregation='cat'
    self.backbone = backbone
    self.relation_head = torch.nn.Sequential(
                             nn.Linear(feature_size, 256),
                             nn.BatchNorm1d(256),
                             nn.LeakyReLU(),
                             nn.Linear(256, 1))

  def aggregate(self, features, K):
    relation_pairs_list = list()
    targets_list = list()
    size = int(features.shape[0] / K)
    shifts_counter=1
    for index_1 in range(0, size*K, size):
      for index_2 in range(index_1+size, size*K, size):
        # Using the 'cat' aggregation function by default
        pos_pair = torch.cat([features[index_1:index_1+size], 
                              features[index_2:index_2+size]], 1)
        # Shuffle by rolling the mini-batch (negatives)
        neg_pair = torch.cat([
                     features[index_1:index_1+size], 
                     torch.roll(features[index_2:index_2+size], 
                     shifts=shifts_counter, dims=0)], 1)
        relation_pairs_list.append(pos_pair)
        relation_pairs_list.append(neg_pair)
        targets_list.append(torch.ones(size, dtype=torch.float32))
        targets_list.append(torch.zeros(size, dtype=torch.float32))
        shifts_counter+=1
        if(shifts_counter>=size): 
            shifts_counter=1 # avoid identity pairs
    relation_pairs = torch.cat(relation_pairs_list, 0)
    targets = torch.cat(targets_list, 0)
    return relation_pairs, targets

  def train(self, tot_epochs, train_loader):
    optimizer = torch.optim.Adam([
                  {'params': self.backbone.parameters()},
                  {'params': self.relation_head.parameters()}])                               
    BCE = torch.nn.BCEWithLogitsLoss()
    self.backbone.train()
    self.relation_head.train()
    for epoch in range(tot_epochs):
      # the real target is discarded (unsupervised)
      for i, (data_augmented, _) in enumerate(train_loader):
        K = len(data_augmented) # tot augmentations
        x = torch.cat(data_augmented, 0)
        optimizer.zero_grad()              
        # forward pass (backbone)
        features = self.backbone(x) 
        # aggregation function
        relation_pairs, targets = self.aggregate(features, K)
        # forward pass (relation head)
        score = self.relation_head(relation_pairs).squeeze()        
        # cross-entropy loss and backward
        loss = BCE(score, targets)
        loss.backward()
        optimizer.step()            
        # estimate the accuracy
        predicted = torch.round(torch.sigmoid(score))
        correct = predicted.eq(targets.view_as(predicted)).sum()
        accuracy = (100.0 * correct / float(len(targets)))
        
        if(i%100==0):
          print('Epoch [{}][{}/{}] loss: {:.5f}; accuracy: {:.2f}%' \
            .format(epoch+1, i+1, len(train_loader)+1, 
                    loss.item(), accuracy.item()))
\end{lstlisting}

\subsection{Main}

\begin{lstlisting}[language=Python,numbers=none]
def main():
  backbone = Conv4() # it should be a CNN with 64 linear output units
  model = RelationalReasoning(backbone)    
  train_set = MultiCIFAR10(K=4, # it should be K=32 for CIFAR10/100
                           root='data', train=True, 
                           transform=train_transform, 
                           download=True)
  train_loader = torch.utils.data.DataLoader(train_set, 
                                             batch_size=64, 
                                             shuffle=True)                                                   
  model.train(tot_epochs=200, train_loader=train_loader)
  torch.save(model.backbone.state_dict(), './backbone.tar')
\end{lstlisting}

\end{document}

%% file: code/code_style.tex
%it requires the packages: xcolor, listings

%New colors defined below
\definecolor{codedarkblue}{rgb}{0,0,0.8}
\definecolor{codegreen}{rgb}{0,0.6,0}
\definecolor{codegray}{rgb}{0.5,0.5,0.5}
\definecolor{codepurple}{rgb}{0.58,0,0.82}
\definecolor{backcolour}{rgb}{0.95,0.95,0.92}

%Code listing style named "mystyle"
\lstdefinestyle{mystyle}{
  backgroundcolor=\color{backcolour},   commentstyle=\color{codegreen},
  keywordstyle=\color{codedarkblue}, %\color{magenta},
  numberstyle=\tiny\color{codegray},
  stringstyle=\color{codepurple},
  basicstyle=\ttfamily\footnotesize,
  breakatwhitespace=false,         
  breaklines=true,                 
  captionpos=b,                    
  keepspaces=true,                 
  numbers=left,                    
  numbersep=5pt,                  
  showspaces=false,                
  showstringspaces=false,
  showtabs=false,                  
  tabsize=2
}

%"mystyle" code listing set
\lstset{style=mystyle}